\documentclass[10pt,twocolumn,letterpaper]{article}

\usepackage{cvpr}
\usepackage{times}
\usepackage{epsfig}
\usepackage{graphicx}
\usepackage{amsmath}
\usepackage{amssymb}

\usepackage[breaklinks=true,bookmarks=false]{hyperref}

\cvprfinalcopy 

\def\cvprPaperID{3854} 

\begin{document}

\title{RF-Net: An End-to-End Image Matching Network based on Receptive Field}

\author{
	\begin{tabular}{cccc}
		Xuelun Shen$^1$ & Cheng Wang$^1$\thanks{Corresponding author.} & Xin Li$^2$ & Zenglei Yu$^1$\\
		Jonathan Li$^{1,3}$ & Chenglu Wen$^1$ & Ming Cheng$^1$ & Zijian He$^1$\\
	\end{tabular}
	\\
	$^1$Fujian Key Laboratory of Sensing and Computing for Smart City,\\
	School of Information Science and Engineering, Xiamen University, China\\
	$^2$School of Electrical Engineering and Computer Science, Louisiana State University, USA\\
	$^3$Department of Geography and Environmental Management, University of Waterloo, Canada\\
	{\tt\small 
		\{cwang,junli,clwen,chm99\}@xmu.edu.cn,
		xinli@cct.lsu.edu
	}\\
	{\tt\small 
		\{xuelun,zengleiyu,kemoho\}@stu.xmu.edu.cn,		
	}
}

\maketitle

\begin{abstract}
This paper proposes a new end-to-end trainable matching network based on receptive field, RF-Net, to compute sparse correspondence between images. Building end-to-end trainable matching framework is desirable and challenging. The very recent approach, LF-Net, successfully embeds the entire feature extraction pipeline into a jointly trainable pipeline, and produces the state-of-the-art matching results. This paper introduces two modifications to the structure of LF-Net. First, we propose to construct receptive feature maps, which lead to more effective keypoint detection. Second, we introduce a general loss function term, neighbor mask, to facilitate training patch selection. This results in improved stability in descriptor training. We trained RF-Net on the open dataset HPatches, and compared it with other methods on multiple benchmark datasets. Experiments show that RF-Net outperforms existing state-of-the-art methods.
\end{abstract}

\section{Introduction}
Establishing correspondences between images plays a key role in many Computer Vision tasks, including but not limited to wide-baseline stereo, image retrieval, and image matching. 
A typical feature-based matching pipeline consists of two components: detecting keypoints with attributions (scales, orientation), and extracting descriptors. 
Many existing methods focus on building/training keypoint detectors or feature descriptors individually.
However, when integrating these separately optimized subcomponents into a matching pipeline, individual performance gain may not directly add up~\cite{Yi2016LIFTLI}.  
Jointly training detectors and descriptors to make them optimally cooperate with each other, hence, is more desirable. 
However, training such a network is difficult because the two subcomponents have their individually different objectives to optimize. 
Not many successful end-to-end matching pipelines have been reported in literatures. 
LIFT~\cite{Yi2016LIFTLI} is probably the first notable design towards this goal. 
However, LIFT relies on the output of SIFT detector to initialize the training, and hence, its detector behaves similarly to the SIFT detector. 
The recent network, SuperPoint~\cite{DeTone2017SuperPointSI}, achieves this end-to-end training. 
But its detector needs to be pre-trained on synthetic image sets, and whole network is trained using images under synthesized affine transformations. 
The more recent LF-Net~\cite{Ono2018LFNetLL} is inspired by Q-learning, and uses a Siamese architecture to train the entire network without the help of any hand-craft method. 
In this paper, we develop an end-to-end matching network with enhanced detector and descriptor training modules, which we elaborate as follows. 

\begin{figure*}[h!bt]	
	\centering
	\begin{tabular}{cc}
		\includegraphics[height=0.18\textheight]{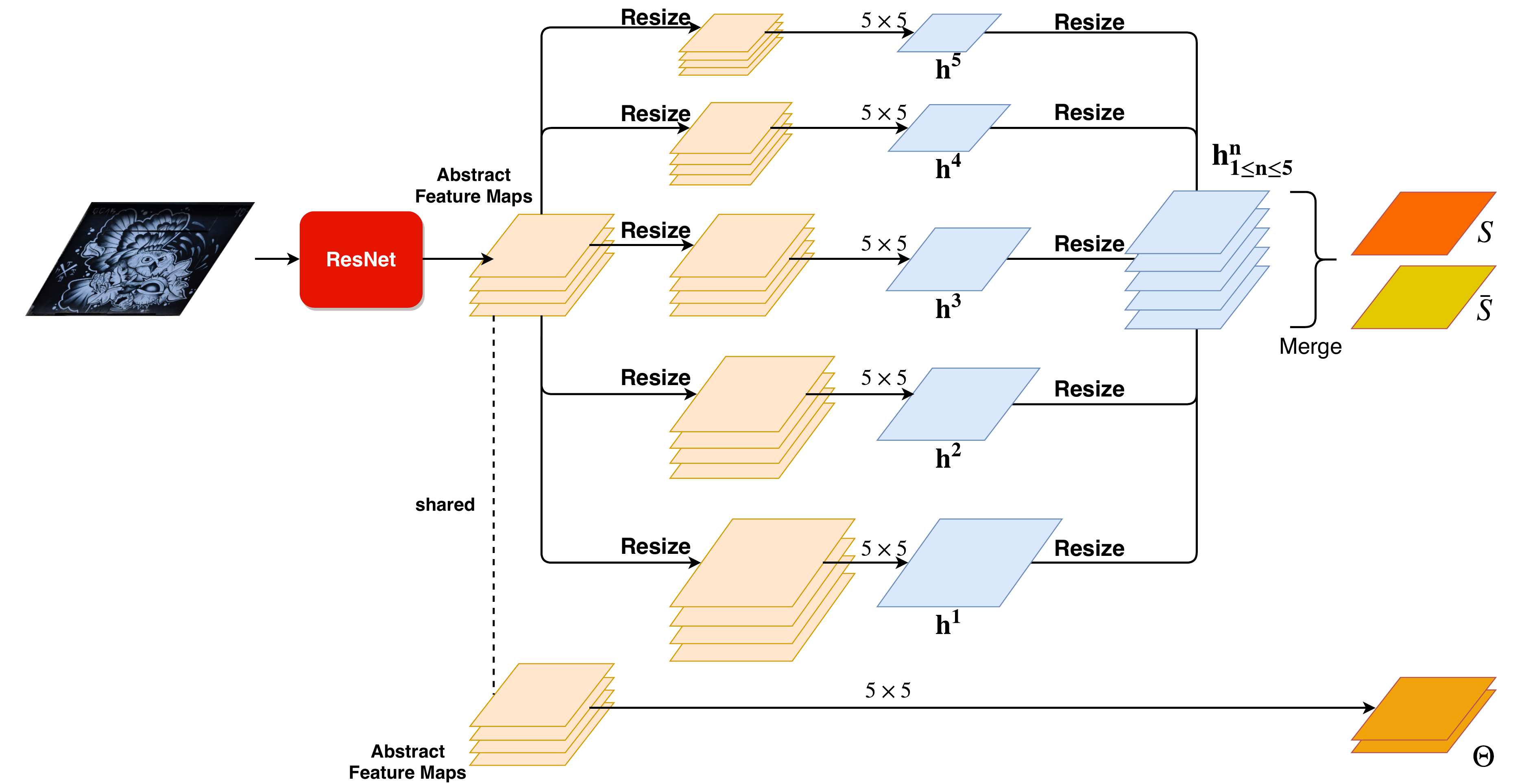}	&
		\includegraphics[height=0.18\textheight]{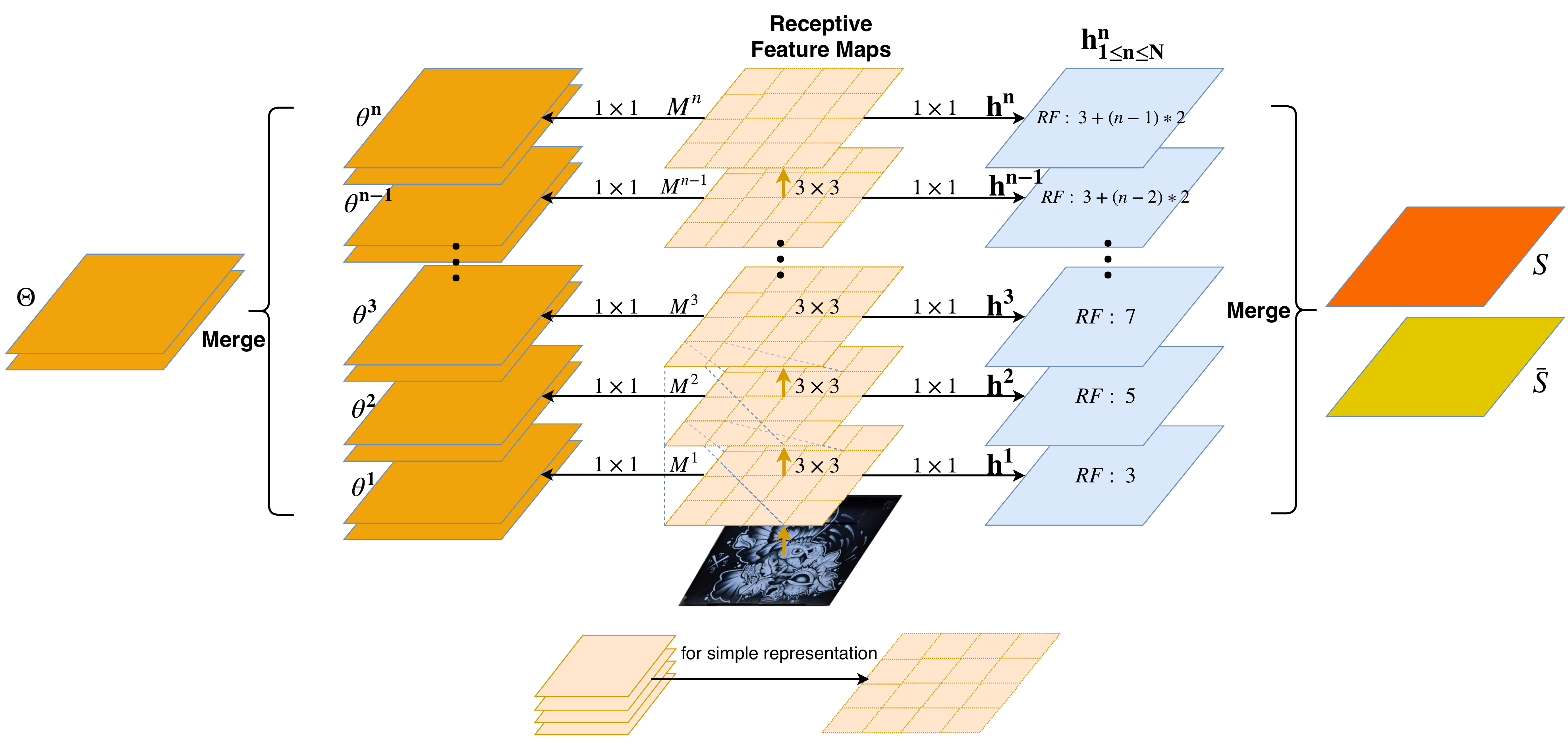} \\	
		(a) LF-Det & (b) RF-Det
	\end{tabular}
	\caption{
		Scale-space response maps $ \mathbf{h}^{n}_{1\leq n\leq N} $ construction in LF-Det (detector in LF-Net~\cite{Ono2018LFNetLL}) and RF-Det (detector in our RF-Net). 
		(a) LF-Det constructs response maps using abstract feature maps extracted from ResNet~\cite{He2016DeepRL}. 
		(b) Our RF-Det constructs response maps using receptive feature maps.
		Note that RF on $ \mathbf{h}^{n} $ represents the receptive field size.
	}
	\label{fig:RFvsLF}
\end{figure*}

\textbf{Keypoint Detection.}
Constructing response maps is a general way to find keypoints. LIFT~\cite{Yi2016LIFTLI} obtains response maps by directly applying convolutions on different resolutions of the input image. 
SuperPoint~\cite{DeTone2017SuperPointSI} does not build response maps, but it processes input image using some convolution and max-pooling layers to produce an intermediate tensor $\mathcal{B}$ whose width and height are only $ \frac{1}{8} $ of the input. 
Hence the response on $ \mathcal{B} $ represents a highly abstract feature of the input image and the size of the feature's receptive field is larger than 8 pixels.  
LF-Net uses ResNet~\cite{He2016DeepRL} to produce abstract feature maps from the input image, then build response maps by convolution on the abstract feature maps at different resolutions.
Therefore, the response on each map has a large receptive field.
In this work, we build response maps using \emph{concerned receptive fields}. 
Specifically, we apply convolution to produce feature maps related to the increasing receptive field (Figure~\ref{fig:RFvsLF} (b)). 
For example, applying convolution with a kernel size of $3\times3$ and stride of 1, the receptive field will increase to 3, 5, 7 and so on. 
This design produces more effective response maps for keypoints detection. 

\textbf{Feature Descriptor.}
Training \emph{descriptors} in an end-to-end network is very different from training individual ones. 
Existing (individual) descriptor training is often done on well-prepared datasets such as the \emph{Oxford} Dataset~\cite{Mikolajczyk2003APE}, UBC \emph{PhotoTour} Dataset~\cite{Winder2007LearningLI}, and \emph{HPatches} Dataset~\cite{balntas2017hpatches}.
In contrast, in the end-to-end network training, patches need to be produced from scratch. 
In LF-Net, patch pairs are sampled by rigidly transforming patches surrounding keypoints in image $I_i$ to image $I_j$.
However, a defect of this simple sampling strategy could affect the descriptor training.
Specifically, two originally far-away keypoints, after transformed, could become very close to each other. 
As a result, a negative patch could look very similar to an anchor patch and positive patch. This will confuse the network during training.
This situation brings labeling ambiguity and effect descriptor training.
We propose a general loss function term called \emph{neighbor mask} to overcome this issue. 
\emph{Neighbor mask} can be used in both \emph{triplet loss} and its variants.

Integrating our new backbone detector and the descriptor network, our sparse matching pipeline is also trained in an end-to-end manner, without involving any hand-designed component. 
We observe that the descriptor's performance greatly influences the detector's training, and a more robust descriptor helps detector learn better. 
Therefore, in each training iteration, we train descriptor twice and detector once. 
To show the effectiveness of our approach through comprehensive and fair evaluations, 
we compare our RF-Net with other methods with three evaluation protocols in two public datasets, \emph{HPatches}~\cite{balntas2017hpatches} and \emph{EF} Dataset~\cite{Zitnick2011EdgeFI}. 
Matching experiments demonstrate that our RF-Net outperforms existing state-of-the-art approaches. 

The main contributions of this paper are in three aspects. 
(1) We propose a new receptive field based detector, which generates more effective scale space and response maps.
(2) We propose a general loss function term for descriptor learning which improves the robustness of patch sampling. 
(3) Our integrated RF-Net supports effective end-to-end training, which leads to better matching performance than existing approaches.

\section{Related work}
A typical feature-based matching pipeline consists of two components: detecting keypoints with attributions (scales, orientation), and extracting descriptors. 
Many recent learning based pipelines focus on improving one of these modules, such as feature detection~\cite{Savinov2017QuadNetworksUL,Zhang2017LearningDA,Rosten2010FasterAB,Verdie2015TILDEAT}, orientation estimation~\cite{Yi2016LearningTA} and descriptor representation~\cite{Mishchuk2017WorkingHT,Tian2017L2NetDL,He2018LocalDO}.
The deficiency of these approaches is that the performance gain from one improved component may not directly correspond to the improvement of the entire pipeline~\cite{Yi2016LIFTLI,Schnberger2017ComparativeEO}.

\textbf{Hand-crafted} approaches like SIFT~\cite{lowe2004distinctive}, is probably the most well-known traditional local feature descriptor. 
A big limitation of SIFT is the speed. SURF~\cite{Bay2008SpeededUpRF} approximates LoG use a box filter and significantly speeds up the detection.
Other popular hand-crafted features include WADE~\cite{Salti2013KeypointsFS}, Edge Foci~\cite{Zitnick2011EdgeFI}, Harris corners~\cite{Harris1988ACC} and its affine-covariant~\cite{Mikolajczyk2004ScaleA}.  


Many effective \textbf{machine-learned detectors} have also been proposed recently. 
FAST~\cite{Rosten2010FasterAB} and ORB~\cite{Rublee2011ORBAE} use machine learning approach to speed up the process of corner detection. 
TILDE~\cite{Verdie2015TILDEAT} learns from pre-aligned images of the same scene at different illumination conditions. 
Although being trained with the assistance from SIFT, TILDE can still identify keypoints missed by SIFT, and perform better than SIFT on the evaluated datasets.
Quad-Network~\cite{Savinov2017QuadNetworksUL} is trained unsupervisedly with a ``ranking'' loss. 
\cite{Zhang2018LearningTD} combines this ``ranking'' loss with a ``Peakedness'' loss and produces a more repeatable detector.
Lenc~\emph{et al.}~\cite{lenc2016learning} proposes to train a feature detector directly from the covariant constraint.
Zhang~\emph{et al.}~\cite{Zhang2017LearningDA} extends the covariant constraint by defining the concepts of ``standard patch" and ``canonical feature".
The method of~\cite{Yi2016LearningTA} learns to estimate orientation to improve feature point matching.

\textbf{Descriptor learning} is the focus of many work for image alignment. 
DeepDesc~\cite{Wang2015ActionRW} applies a Siamese network, MatchNet~\cite{Han2015MatchNetUF} and Deepcompare~\cite{Zagoruyko2015LearningTC}, to learn nonlinear distance matrix for matching. 
A series of recent works have considered more advanced model architectures and triplet-based deep metric learning formulations, including UCN~\cite{Choy2016UniversalCN}, TFeat~\cite{Balntas2016LearningLF}, GLoss~\cite{Kumar2016LearningLI}, L2-Net~\cite{Tian2017L2NetDL}, Hard-Net~\cite{Mishchuk2017WorkingHT} and He~\emph{et al.}~\cite{He2018LocalDO}.
Recent works focus on designing better loss functions, while still using the same network architecture proposed in L2-Net~\cite{Tian2017L2NetDL}.

Building \textbf{end-to-end matching frameworks} have been less explored.
LIFT~\cite{Yi2016LIFTLI} was probably the first attempt to build such a network. 
It combines three CNNs (for the detector, orientation estimator, and descriptor) through differentiable operations.
While it aims to extract an SfM-surviving subset of DoG detections, its detector and orientation estimator are fed with a patch instead of the whole image, and hence, are not trained end-to-end.
SuperPoint~\cite{DeTone2017SuperPointSI} trains a fully-convolutional neural network that consists of a single shared encoder and two separate decoders (for feature detection and description respectively). 
Synthetic shapes are used to generate images for detector's pre-training, and synthetic homographic transformations are used to produce image pairs for detector's fine-tuning.
The more recent LF-Net~\cite{Ono2018LFNetLL} presents a novel deep architecture and a training strategy to learn a local feature pipeline from scratch. Based on a Siamese Network structure, LF-Net predicts on one branch, and generates ground truth on another branch. 
It is fed with a QVGA sized image and produces multi-scale response maps. Next, it processes the response maps to output three dense maps, representing keypoints saliency, scale, and orientation, respectively.

\begin{figure*}[h!tb]	
	\centering
	\includegraphics[width=\textwidth]{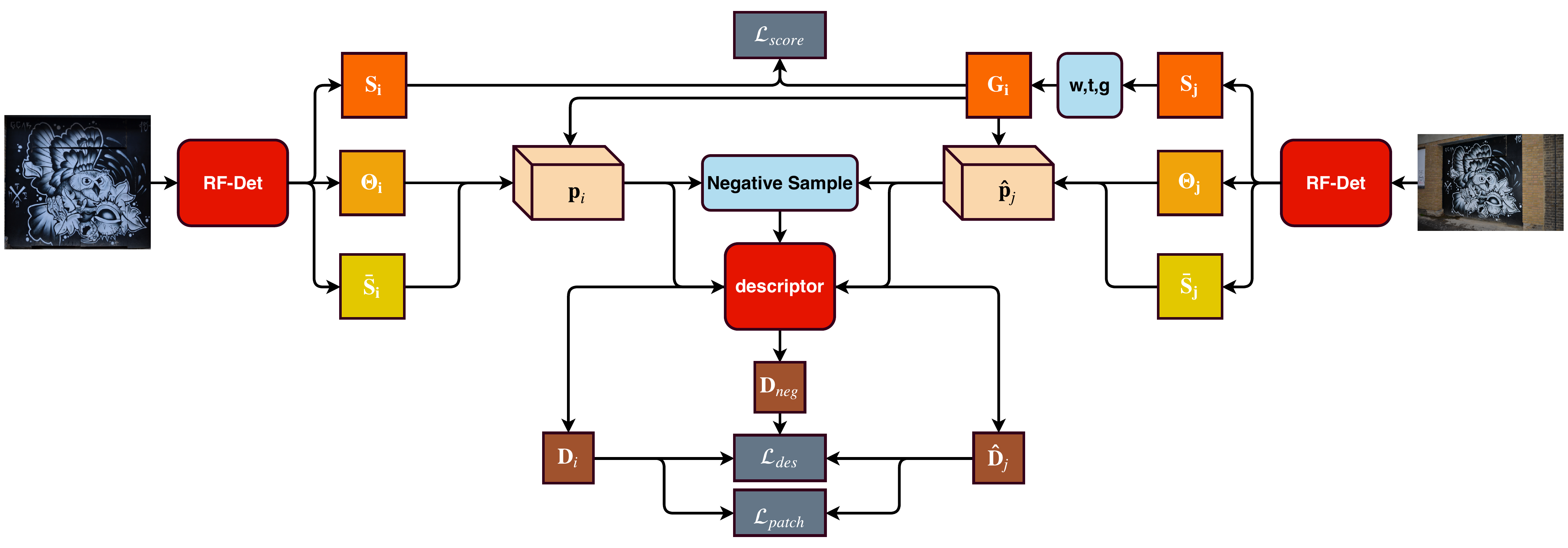}	
	\caption{
		The whole network structure for RF-Net. In training, we feed one pair of images into the network. 
		The image on the right is processed by the network to generate the ground truth of the left image.
		After calculating the gradient of the loss function, parameters are updated by back propagation.
		Next we exchange the positions of the two images and train the network again.
	}
	\label{fig:networkstructure}
\end{figure*}

\section{Approach}
Our RF-Net consists of a detector, called RF-Det, which is based on receptive feature maps, and a description extractor
whose architecture is the same as L2-Net~\cite{Tian2017L2NetDL}, but with a modified loss function.
The design of the whole network structure is depicted in Figure~\ref{fig:networkstructure}.
During testing, the detector network RF-Det takes in an image and outputs a score map $\mathbf{S}$, an orientation map $\mathbf{\Theta}$, and a scale map $\mathbf{\bar{S}}$.
These three maps produce the locations, orientations, and scales of keypoints, respectively.
Patches cropped from these maps will be fed to the descriptor module to extract fixed-length feature vectors for matching.

\subsection{Scale Space Response Maps Construction}
Constructing scale space response maps is the basis for keypoint detection.
We denote the response maps as $\{ \mathbf{h}^n \}$, where $1\leq n\leq N$ and $N$ is the total layer number.
The LF-Net~\cite{Ono2018LFNetLL} uses abstract feature maps extracted from ResNet~\cite{He2016DeepRL} to construct its response maps. 
Each response in the abstract feature maps represents a high-level feature extracted from a large region in the image, while the low-level features are not extracted.
Thus, every map in $ \mathbf{h}^{n}$ is a large-scale response in the scale space.

Our idea is to preserve both high-level and low-level features when constructing the response maps $\{\mathbf{h}^{n} \} $,
and use some maps (e.g., with smaller index) to offer small-scale response,
and some others (e.g., with bigger index) to offer
large-scale response.

Following this idea, we use $N$ hierarchical convolutional layers to produce feature maps $\{ \mathbf{M}^{n} \}, 1\leq n\leq N$ with increasing receptive fields.
Therefore, each response in $ \mathbf{M}^{n}$ describes the abstracted features extracted from a certain range of the image, and this range increases as the convolution applies.
Then we apply one $ 1\times1 $ convolution on each $ \mathbf{M}^{n}$ to produce response maps $ \mathbf{h}^{n}_{1\leq n\leq N} $ in the multi scale space.

In our implementation, we set $N=10$. And the hierarchical convolutional layers consist of sixteen $3\times3 $ kernels followed by an instance normalization~\cite{Ulyanov2016InstanceNT} and leaky ReLU activations.
We also add shortcut connection~\cite{He2016DeepRL} between each layer, which does not change the receptive field in feature maps and makes training of the network easier.
To produce multi-scale response maps $ \mathbf{h}^{n}$, we use one $1\times1$ kernel followed by an instance normalization. All convolution are zero-padded to make the output size same as the input.

\subsection{Keypoint Detection}
Following the commonly adopted strategy, we select high-response pixels as keypoints.
Response maps $ \mathbf{h}^{n} $ represent pixels' response on multi-scales, so we produce the keypoint score map from it.
Then we design the keypoint detection similar to LF-Net~\cite{Ono2018LFNetLL}, except that our response maps
$\mathbf{h}^{n}$ are constructed by receptive feature maps.

Specifically, we perform two softmax operators to produce the score map $\mathbf{S} $.
The purpose of the first softmax operator is to produce sharper response maps $ \hat{\mathbf{h}}^{n} $.
The first softmax operator is applied over a
$15\times15\times{N} $ window sliding on $ \mathbf{h}^{n}$ with the same zero padding.
Then we merge all the $\hat{ \mathbf{h}}^{n} $ into the final score map $ \mathbf{S} $ with the second $softmax_n$  operator, by
\begin{equation}
Pr^{n} = \mathit{softmax}_{n}(\mathbf{\hat{h}}^{n}),
\end{equation}
and
\begin{equation}
\mathbf{S}=\sum_{n}\mathbf{\hat{h}}^{n}\odot{Pr^{n}},
\end{equation}
where $ \odot $ is the Hadamard product, and $ Pr^{n} $ indicates the probability of a pixel being a keypoint.
The second $softmax_n$ is applied on a $ 1\times1\times{N} $ window sliding on $ \hat{\mathbf{h}}^{n} $.

Estimations of the orientation and scale are also produced based on $ Pr^{n} $.
We apply convolutions on $ \mathbf{M}^{n} $ with two $1\times1$ kernels to produce multi-scale orientation maps
$\{ \mathbf{\theta}^{n} \}$ (see Figure~\ref{fig:RFvsLF} (b)) whose values indicate the $ sine $ and $ cosine $ of the orientation.
The values are used to compute an angle using the $ arctan $ function. Then we apply the same product to merge all $ \mathbf{\theta}^{n} $ into the final orientation map $ \mathbf{\Theta} $, by
\begin{equation}
\mathbf{\Theta}=\sum_{n}\mathbf{\theta}^{n}\odot{Pr^{n}}.
\end{equation}
To produce the scale map $ \mathbf{\bar{S}} $, we apply the similar operation used in orientation estimation:
\begin{equation}
\mathbf{\bar{S}}=\sum_{n}\mathbf{\bar{s}}^{n}\odot{Pr^{n}},
\end{equation}
where $ \mathbf{\bar{s}}^{n} $ is the receptive field size of the $ \mathbf{h}^{n} $.

\subsection{Descriptor extraction}
We develop the descriptor extraction module in the network following a structure similar to the L2-Net~\cite{Tian2017L2NetDL}.
This structure is also adopted in other recent descriptor learning frameworks such as Hard-Net~\cite{Mishchuk2017WorkingHT} and He~\emph{et al.}~\cite{He2018LocalDO}.
Specifically, this descriptor network consists of seven convolution layers, each followed by a batch normalization and ReLU, except for the last one.
The output descriptors are L2 normalized, and its dimension is 128.
We denote the output descriptors as $ \mathbf{D} $.
While we adopt this effective network structure similar to many recent descriptor extraction modules, we use a different loss function, which is discussed in the following.

\subsection{Loss Function}
A keypoint detector predicts keypoints' locations, orientations, and scales. Therefore, its loss function consists of \emph{score loss} and \emph{patch loss}.
Patch descriptor is independent from the detection component, once the keypoints are selected. Hence, we use another \emph{description loss} to train it.

\begin{figure*}[h!bt]
	\centering
	\begin{tabular}{cccc}
		\includegraphics[height=0.11\textheight]{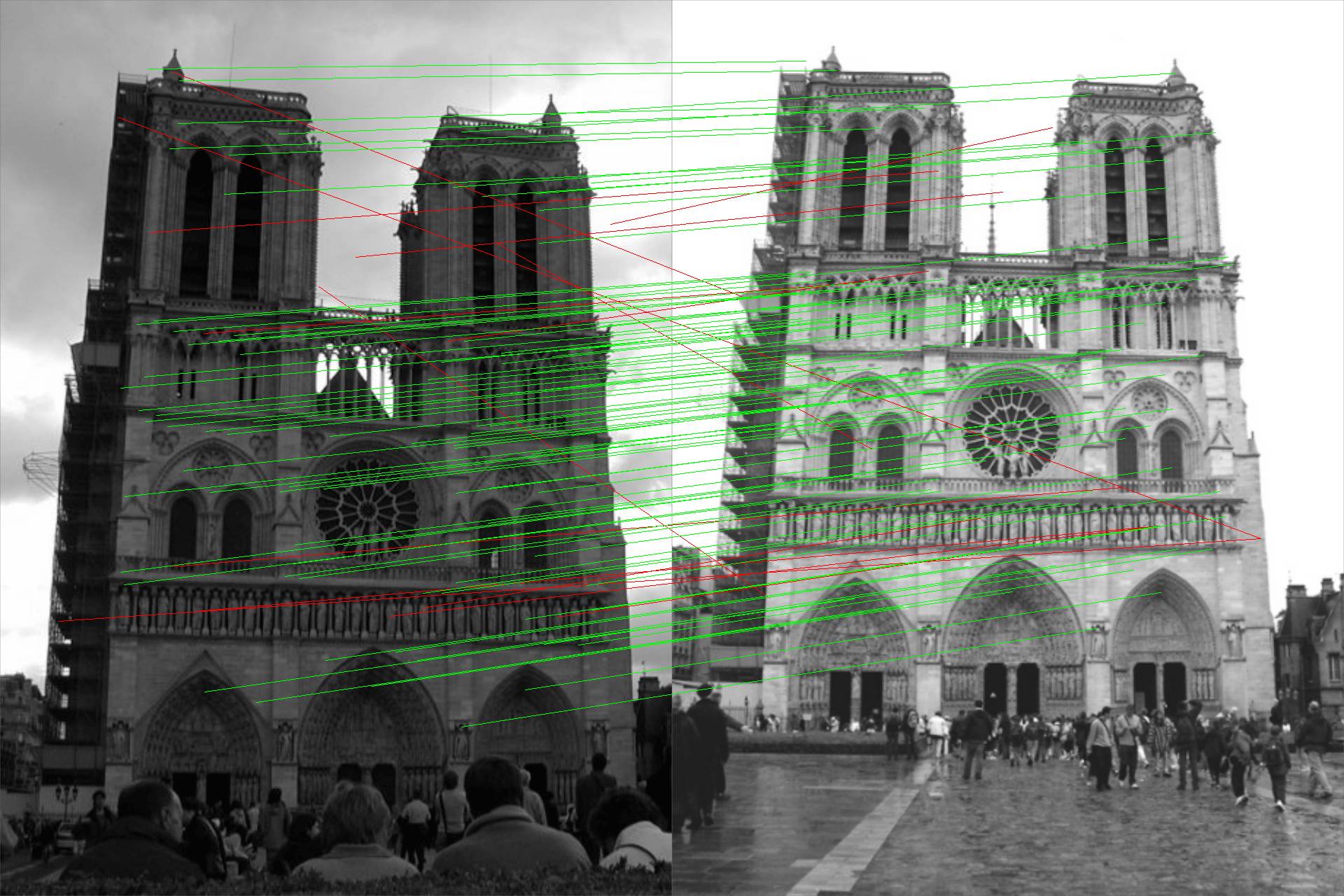} &
		\includegraphics[height=0.11\textheight]{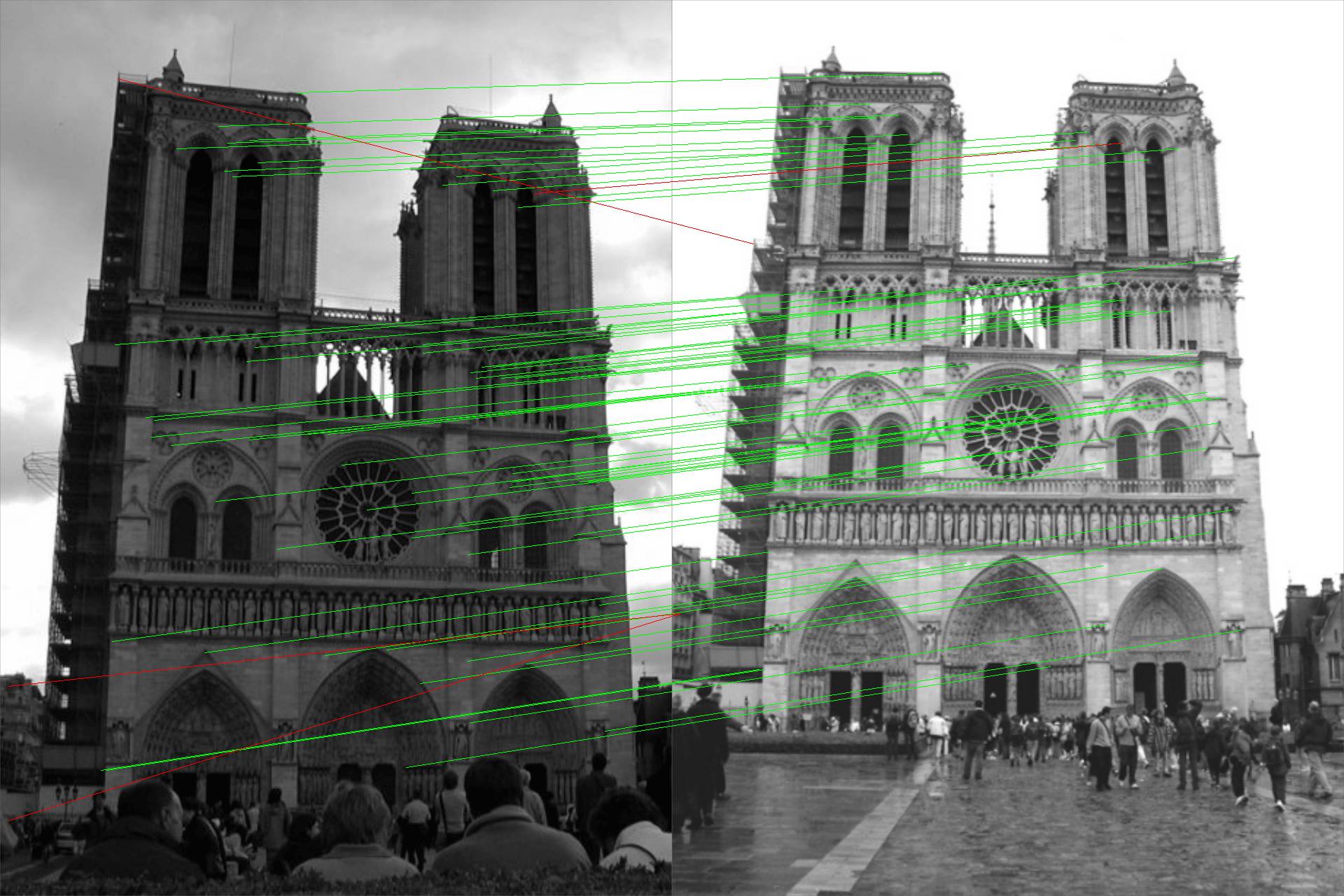} &
		\includegraphics[height=0.11\textheight]{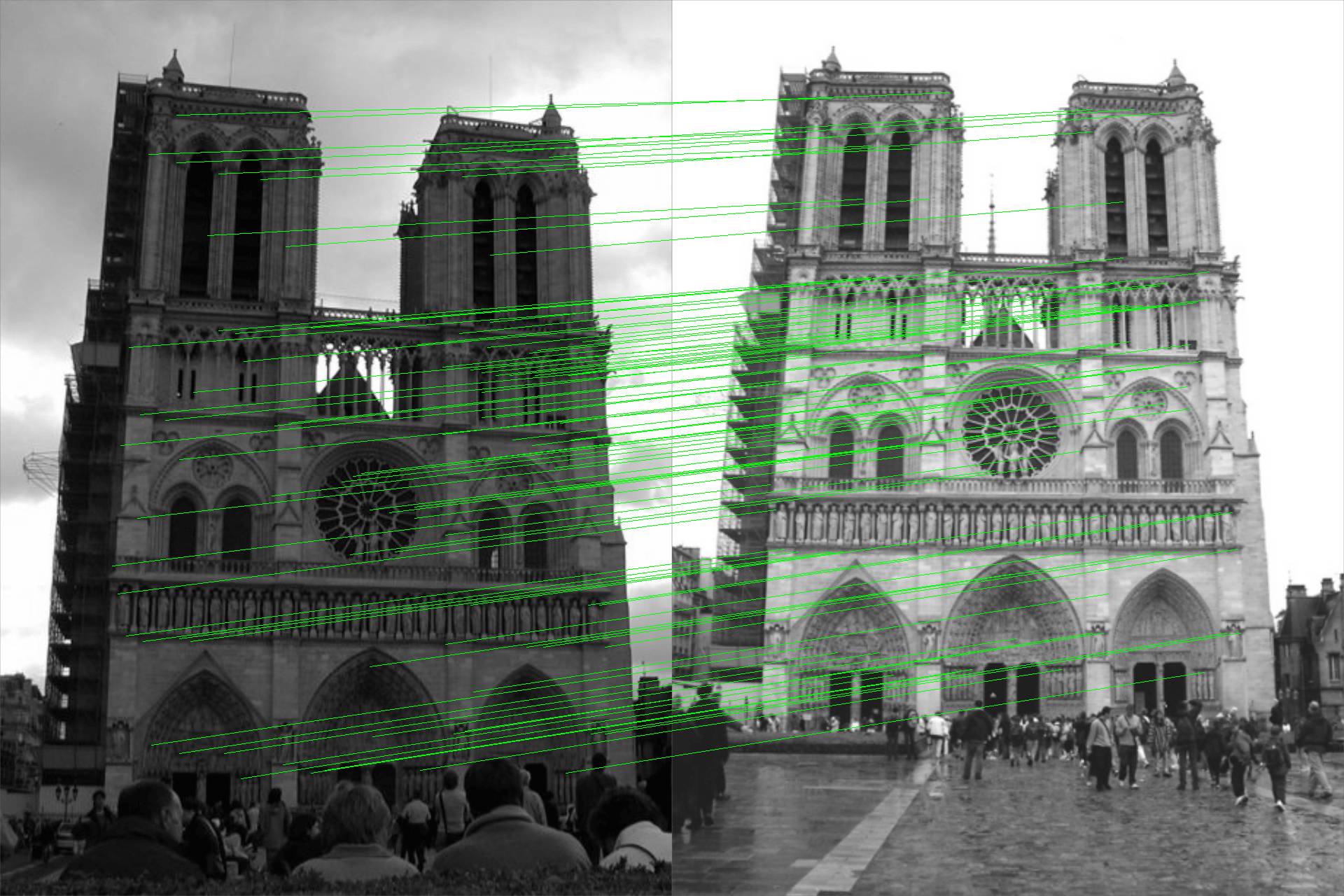}   &
		\includegraphics[height=0.11\textheight]{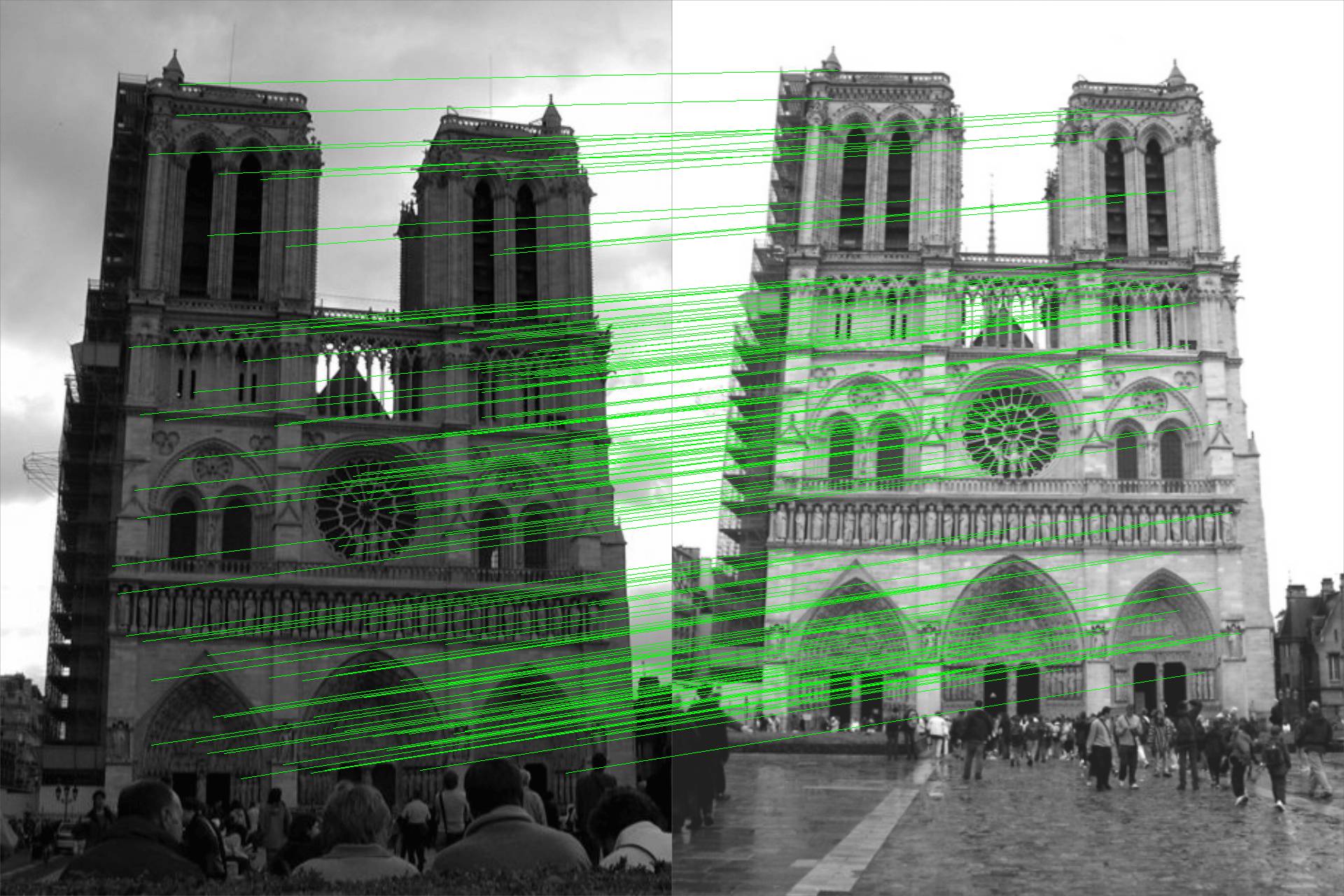}   \\

		\includegraphics[height=0.11\textheight]{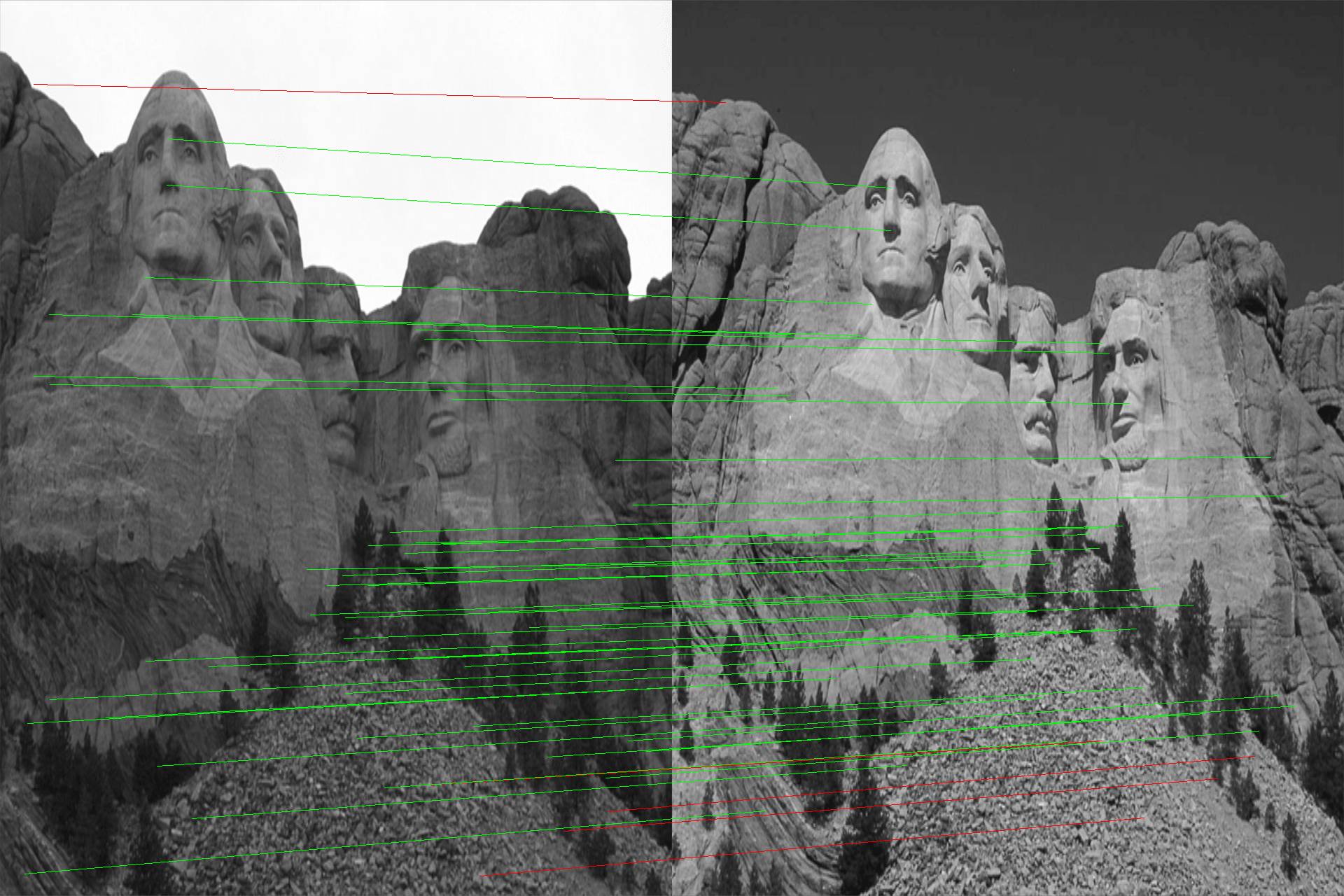} &
		\includegraphics[height=0.11\textheight]{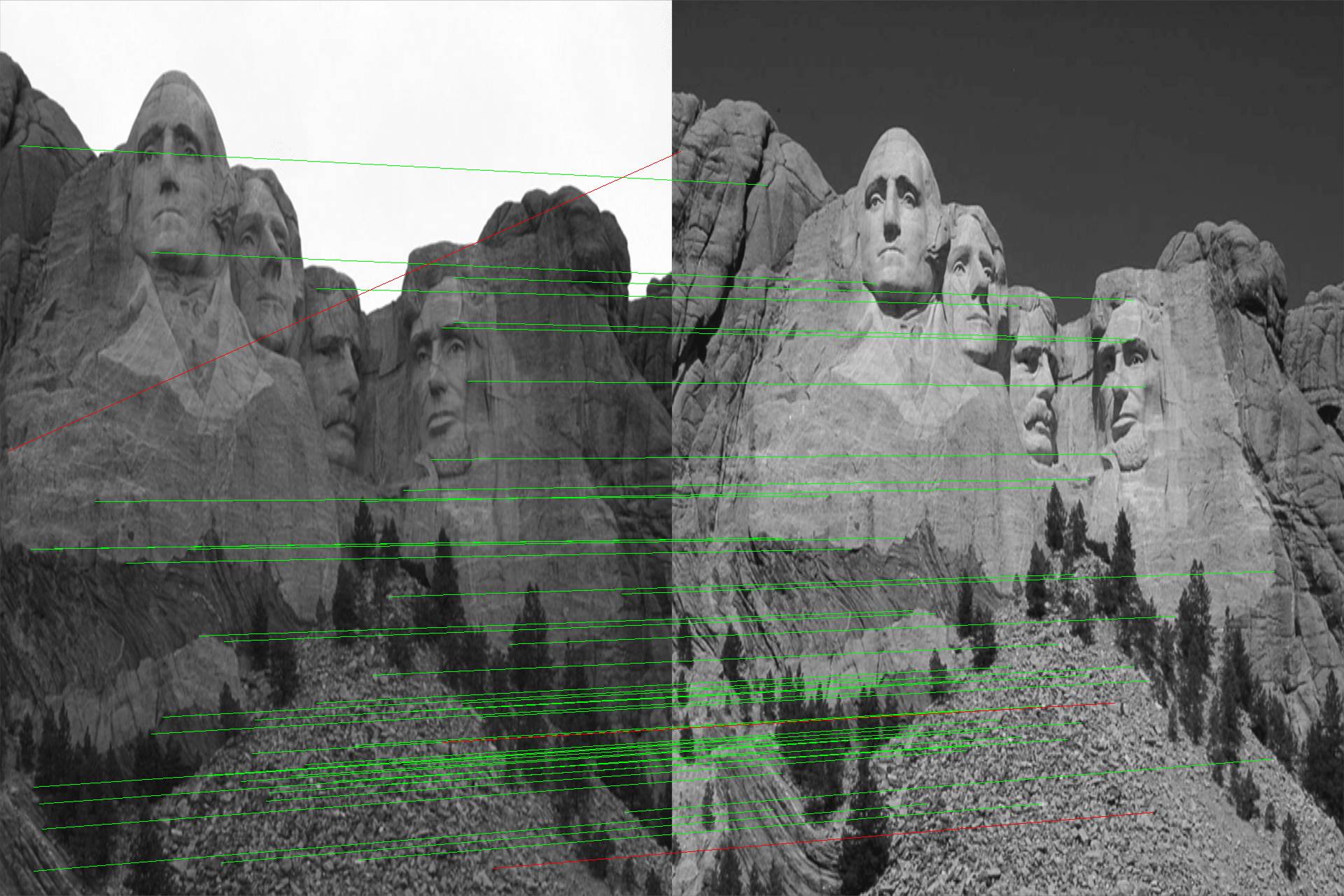} &
		\includegraphics[height=0.11\textheight]{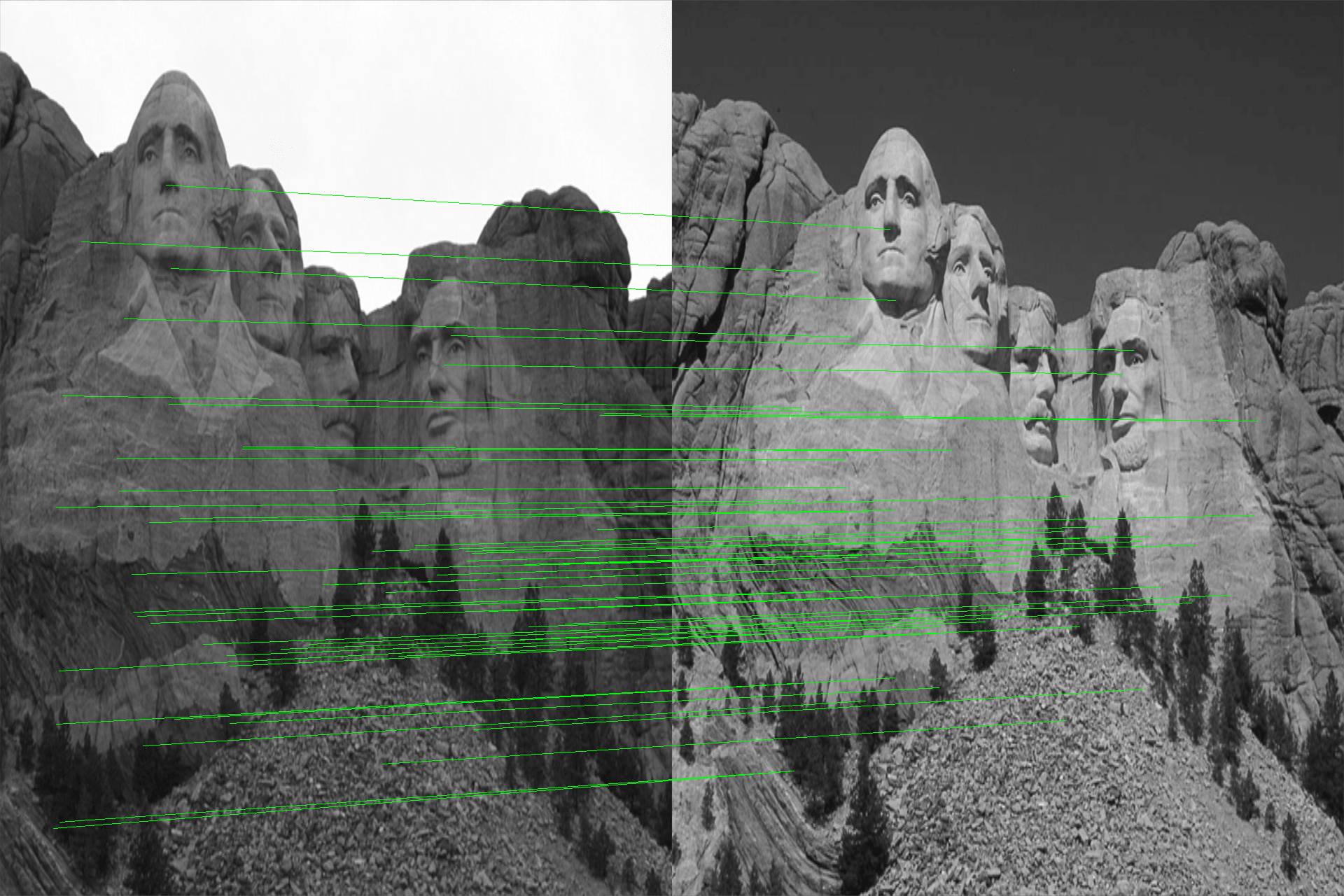}   &
		\includegraphics[height=0.11\textheight]{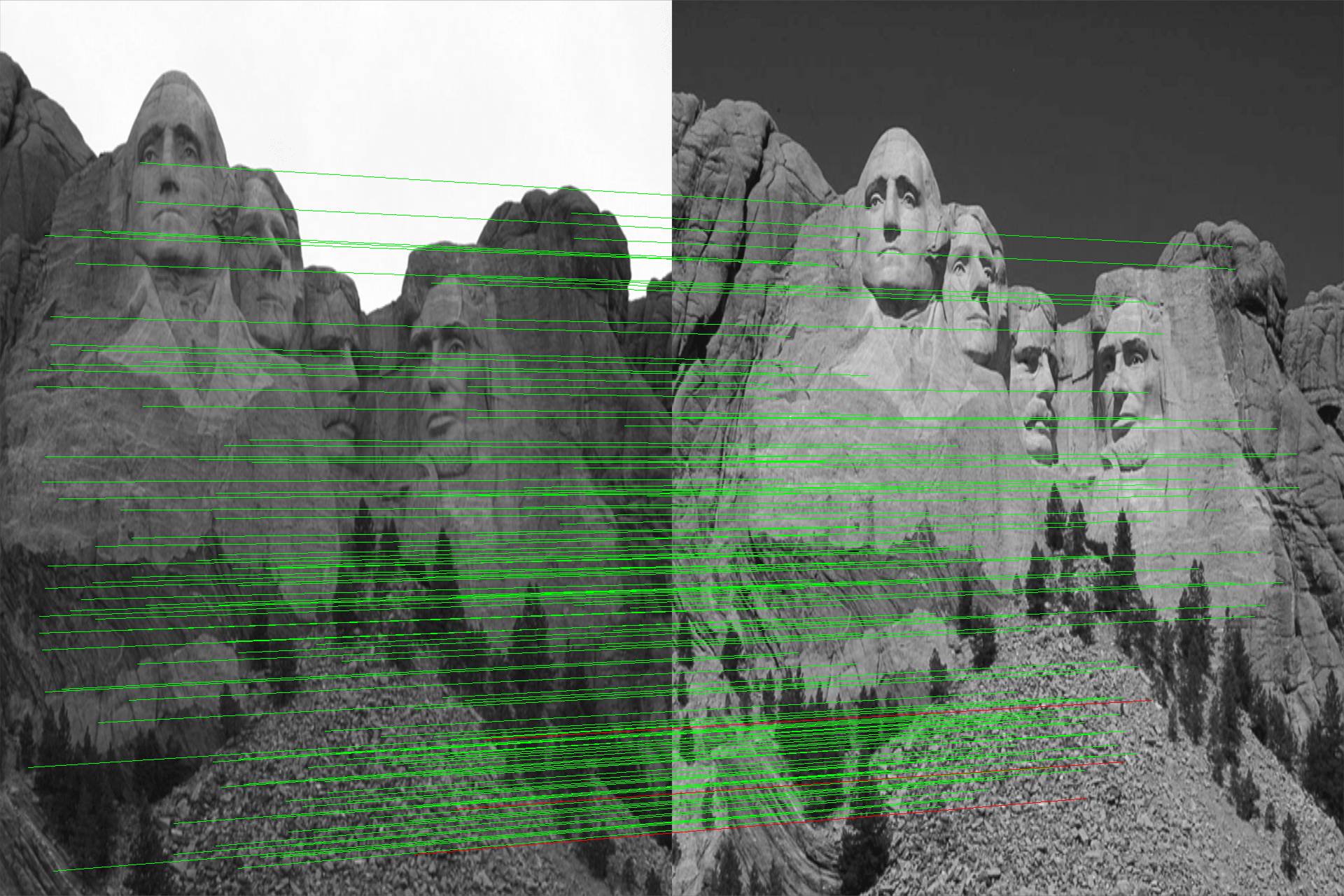}   \\

		\includegraphics[height=0.11\textheight]{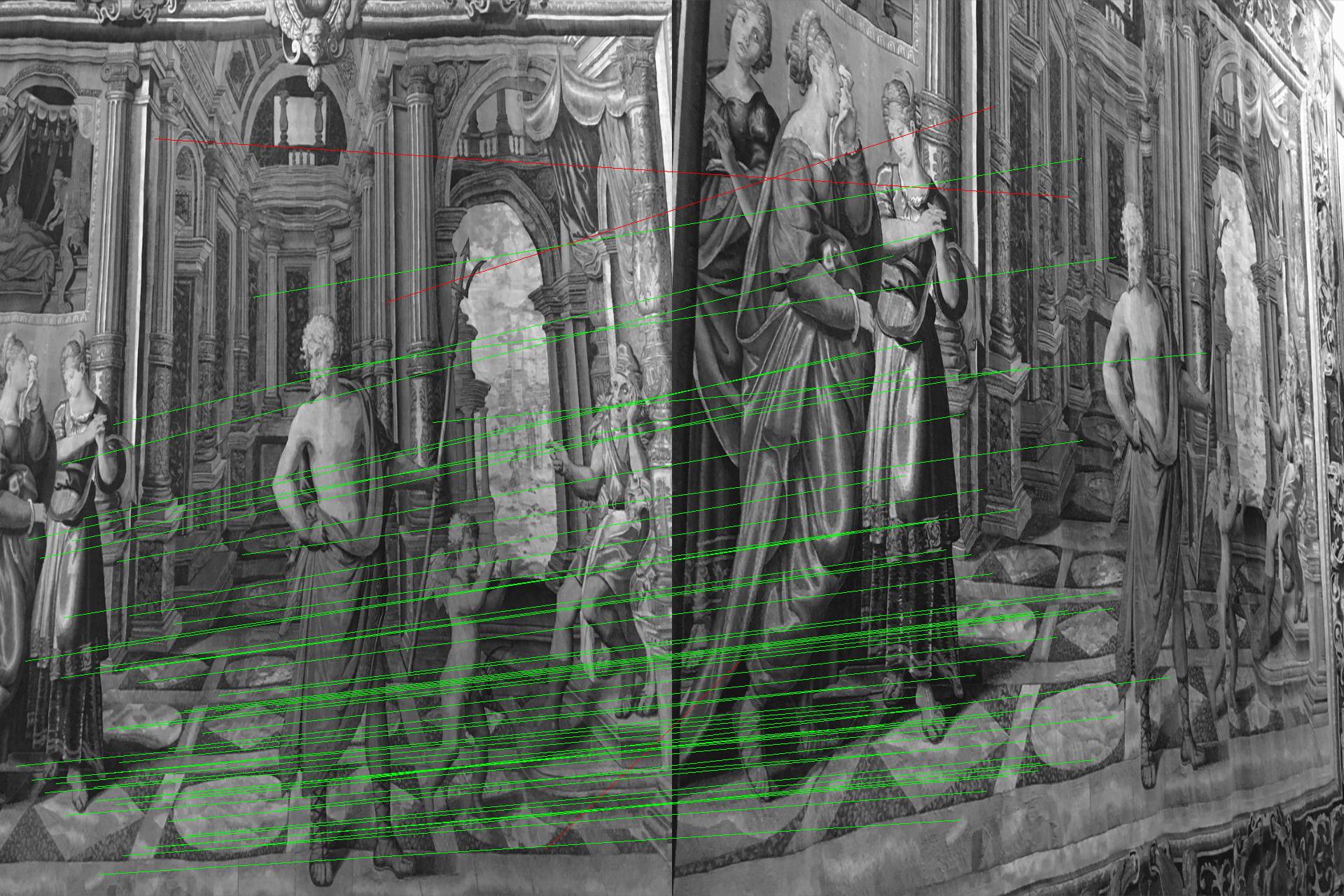} &
		\includegraphics[height=0.11\textheight]{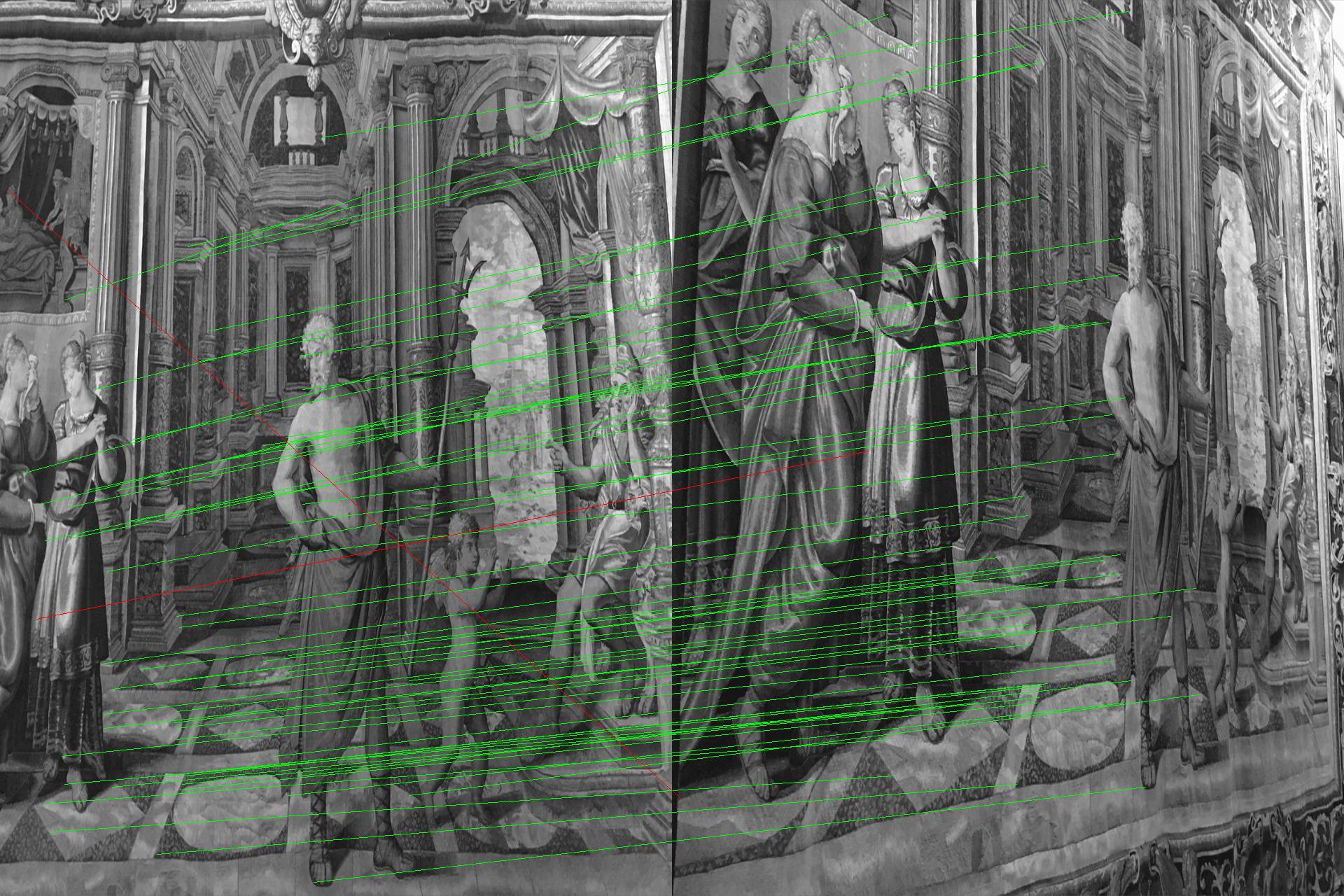} &
		\includegraphics[height=0.11\textheight]{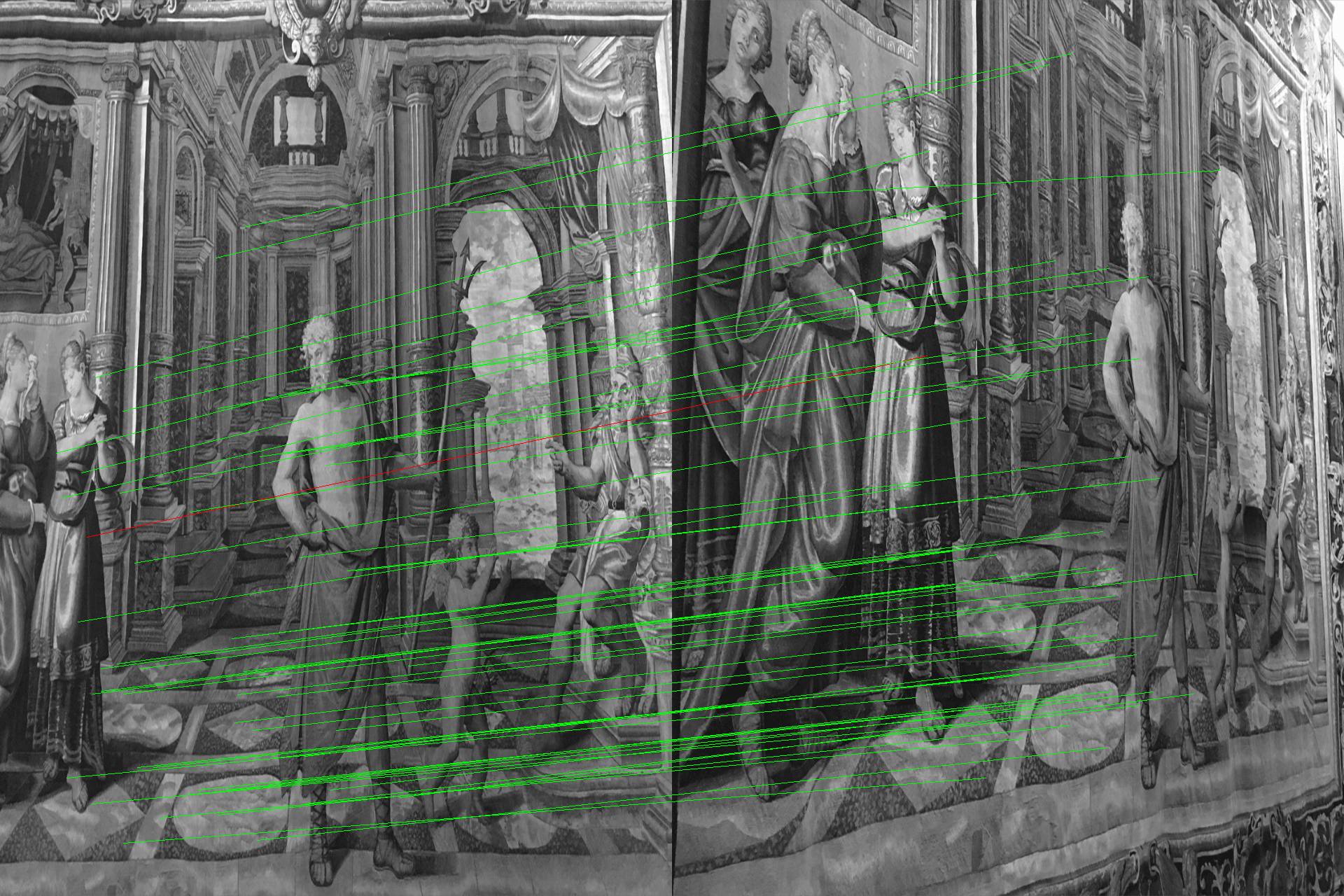}   &
		\includegraphics[height=0.11\textheight]{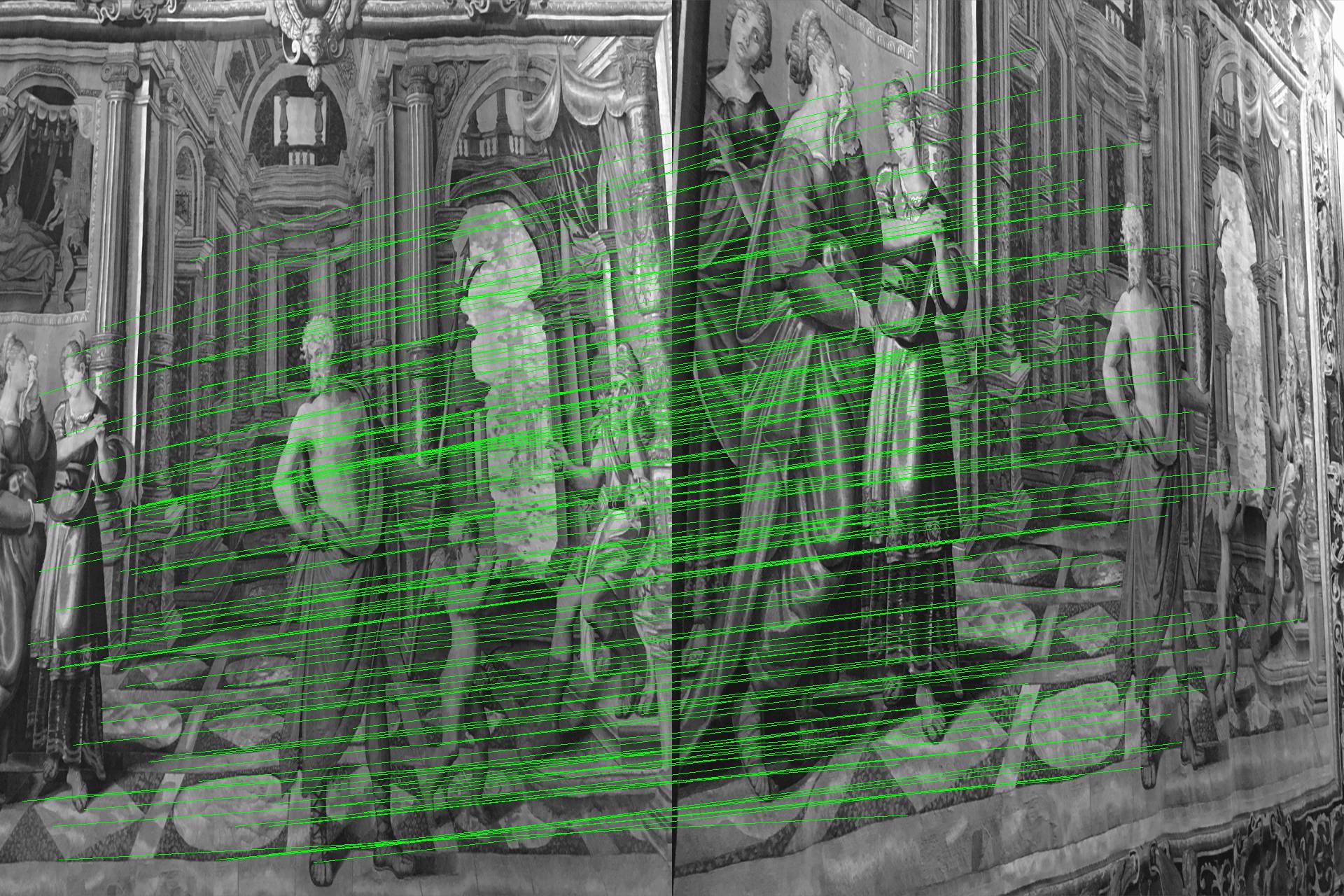}   \\

		\includegraphics[height=0.11\textheight]{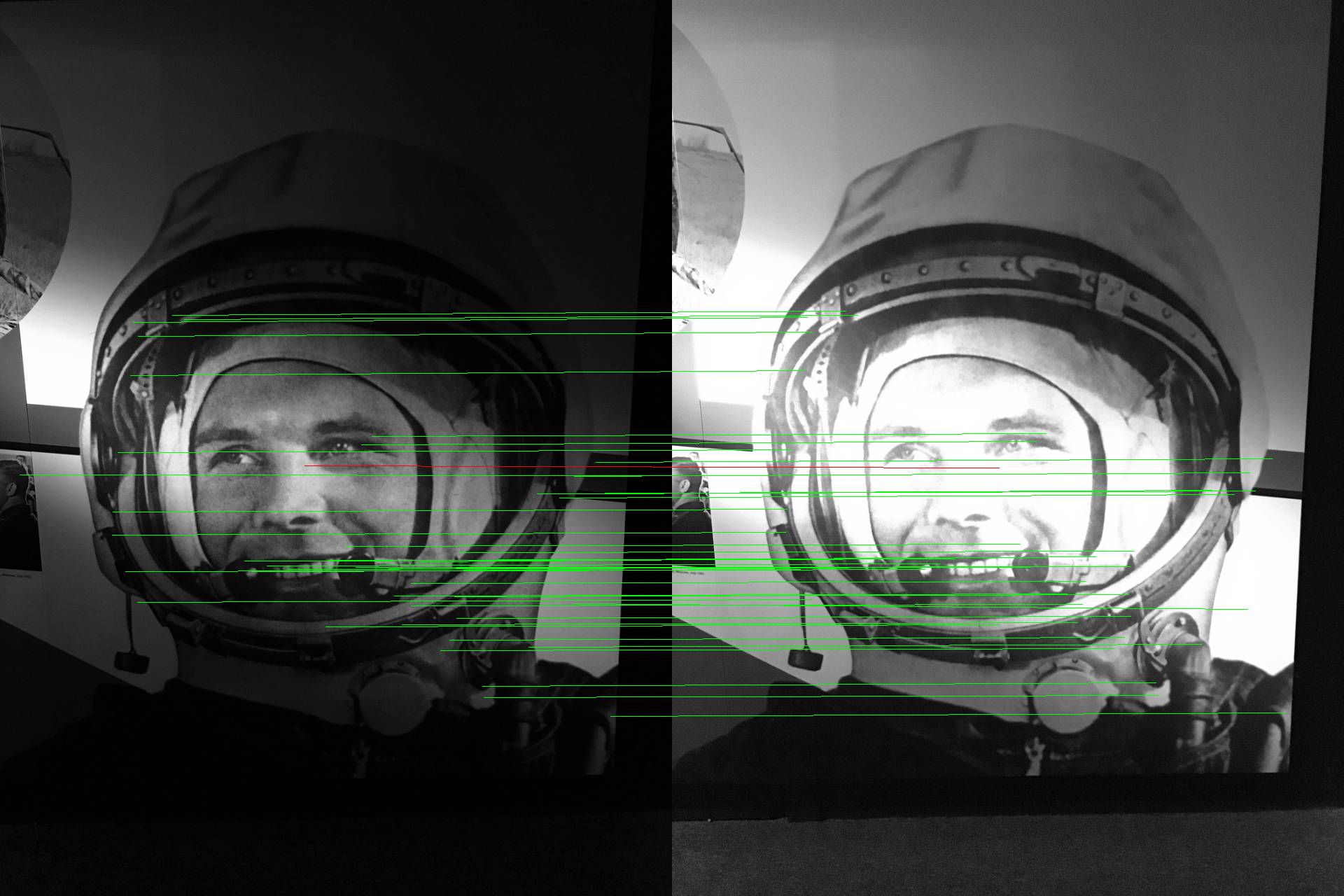} &
		\includegraphics[height=0.11\textheight]{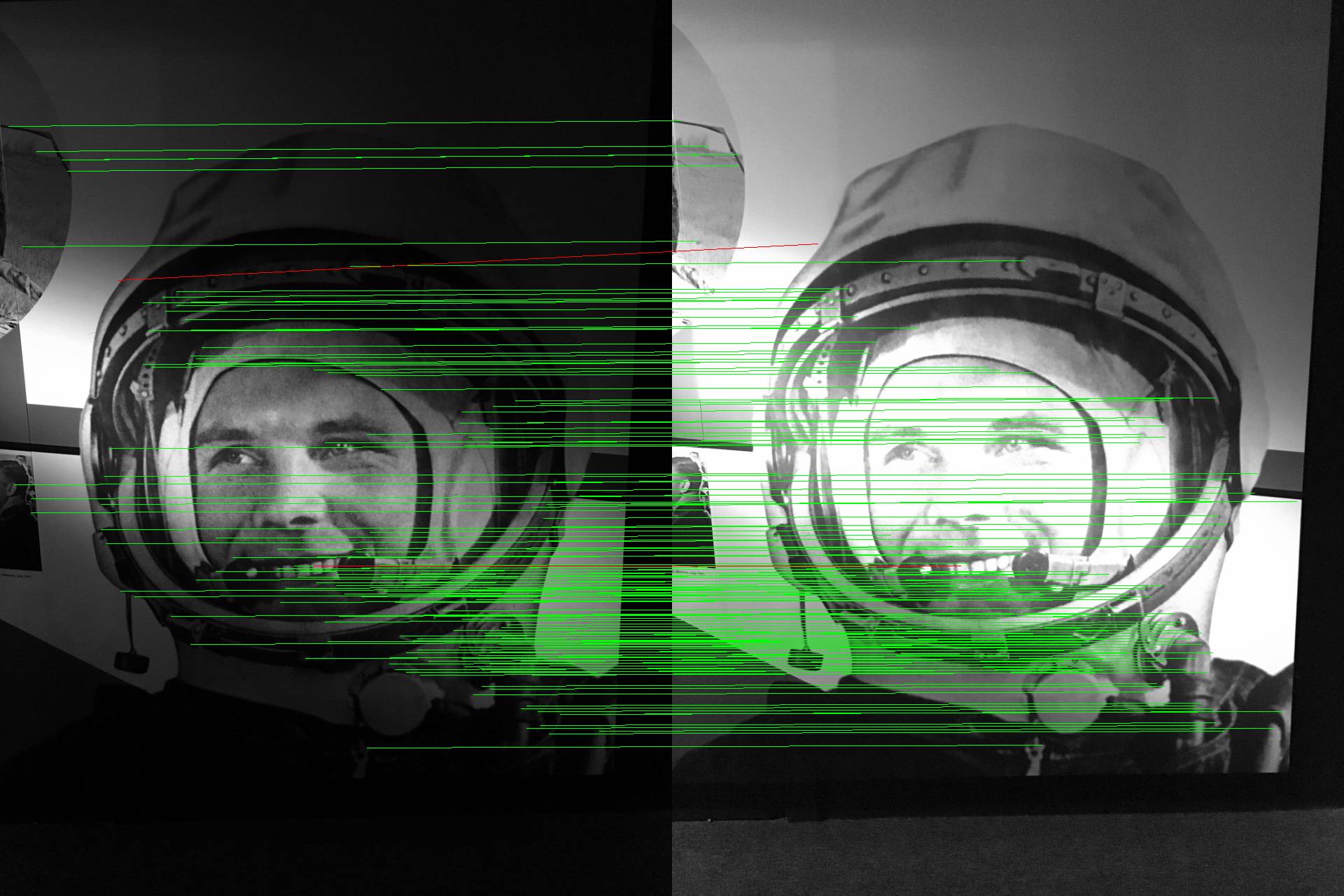} &
		\includegraphics[height=0.11\textheight]{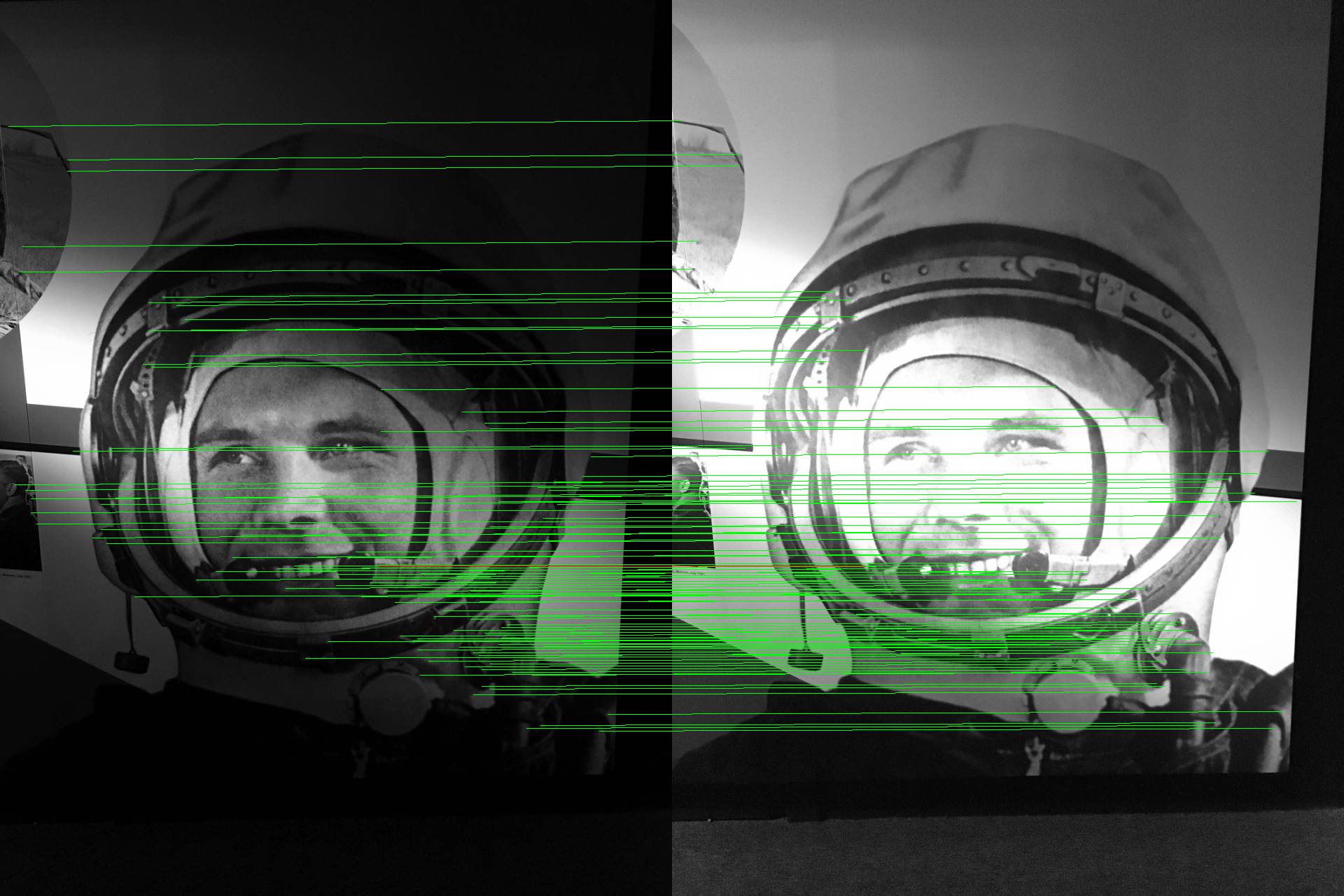}   &
		\includegraphics[height=0.11\textheight]{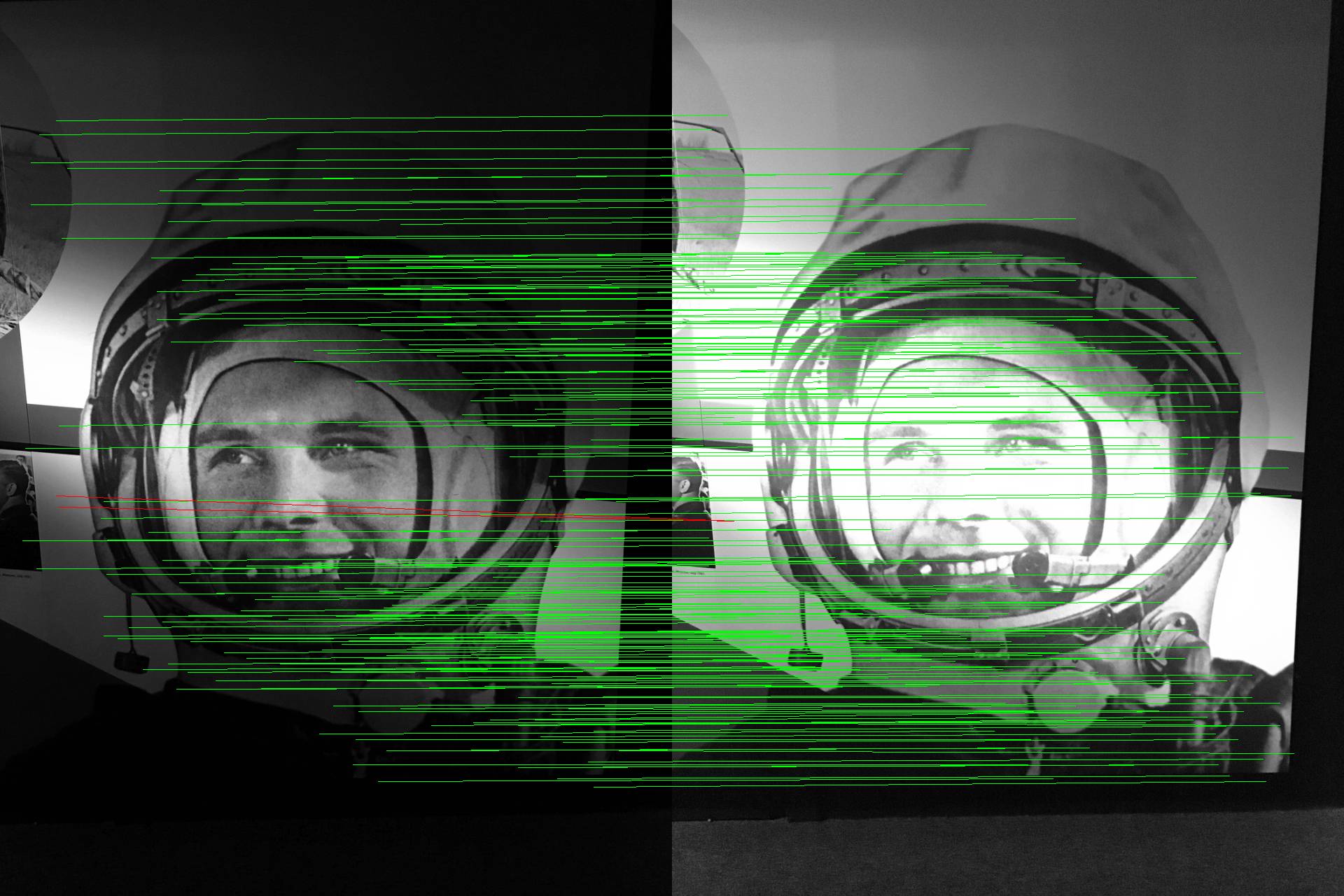}   \\

		(a) SIFT & (b) FAST+Hard-Net & (c) LF-net & (d) RF-Net
	\end{tabular}
	\caption{
		Qualitative matching results, with correct matches drawn in green lines and failed matches drawn in red lines.
		These columns are SIFT, FAST detector integrated with Hard-Net descriptor, LF-Net, and RF-Net.
		The images in top two rows are from \emph{EF} Dataset~\cite{Zitnick2011EdgeFI}, and the images in bottom two rows are from \emph{HPatches}~\cite{balntas2017hpatches}.
		We use the nearest neighbor distance ratio($ 0.7 $) as matching strategy with $K=1024$ keypoints to match two images.
		As the figure showed, more green lines and fewer red lines means better matching results.
	}
	\label{fig:qualitative}
\end{figure*}

\textbf{Score loss.}
In this feature matching problem, because it is unclear which points are important, we cannot produce ground truth score maps through human labeling.
Good detectors should be able to find the corresponding interest points when the image undergoes a transformation.
A simple approach is to let the two score maps $ \mathbf{S}_{i} $ and $ \mathbf{S}_{j} $ (produced from images $ I_{i} $ and  $ I_{j} $, respectively) to have the same score at the corresponding locations.
A simple approach to implement the idea is to minimize Mean Square Loss (MSE) between corresponding locations on $ \mathbf{S}_{i} $ and $ \mathbf{S}_{j} $.
However, this approach turned out to be not very effective in our experiments.

LF-Net suggests another approach. We fed image pair $ I_{i} $ and $ I_{j} $ into network to produce $\mathbf{S}_{i}$ and $\mathbf{S}_{j}$.
We process $\mathbf{S}_{j}$ to produce ground truth $\mathbf{G}_{i}$, then define the score loss to be the MSE between $\mathbf{S}_{i}$ and $\mathbf{G}_{i}$.
More specifically, given the ground truth perspective matrix, first, we select the top $K$ keypoints from the warped score map $\mathbf{S}_{j}$, and we denote this as operation $ \mathit{t} $.
Then, we generate a clean ground truth score map $\mathbf{G}_{i}$ by placing Gaussian kernels with standard deviation $\sigma=0.5$ at those locations. This operation is denoted as $\mathit{g}$.
Then for warping, we apply a perspective transform $\mathit{w}$. This score loss is finally written as:
\begin{equation}
\mathbf{G}_{i} = \mathit{g(\mathit{t}(\mathit{w}(\mathbf{S}_{j})))},
\end{equation}
\begin{equation}
\mathcal{L}_{score}(\mathbf{S}_{i}, \mathbf{S}_{j})=|\mathbf{S}_{i}-\mathbf{G}_{i}|^{2}.
\end{equation}
If a keypoint falls outside the image $I_i$, we drop it from the optimization process.

\textbf{Patch loss.}
Keypoint orientation and scale affect the patches cropped from the image; and descriptors extracted from patches influence matching precision.
We define a \emph{patch loss} to optimize detector to detect more consistent keypoints.
We hope that the patches cropped from the corresponding position are as similar as possible.

Specifically, we select the top $K$ keypoints from $\mathbf{G}_{i}$, then warp their spatial coordinates back to $I_{j}$, and form the keypoint with orientation and scale from $\mathbf{\Theta}$ and $\mathbf{\bar{S}}$ predicted by each image. We extract descriptors $\mathbf{D}^{k}_{i}$ and $\mathbf{\hat{D}}^{k}_{j}$ at these corresponding patches $\mathbf{p}^{k}_{i}$ and $\mathbf{\hat{p}}^{k}_{j}$. The patch loss can be formulated as
\begin{equation}
\mathcal{L}_{patch}(
\mathbf{D}^{k}_{i},
\mathbf{\hat{D}}^{k}_{j})
=
\frac{1}{K}
\sum_{k=1,K}
d(\mathbf{D}^{k}_{i},
\mathbf{\hat{D}}^{k}_{j}
),
\end{equation}
where
\begin{equation}
\mathit{d}(x,y) = \sqrt{2-2xy}.
\end{equation}

Unlike LF-Net that selects keypoints from $\mathbf{I}_{i}$, we select keypoints from $\mathbf{G}_{i}$.
This is because in many public training datasets (e.g. HPatches), there is no background mask available.
After transformed, Keypoints selected from $\mathbf{I}_{i}$ may be out of range on image $\mathbf{I}_{j}$. Therefore, the  training data sampling method we use is more general.

\textbf{Description loss.}
The description loss we use is based on the \emph{hard loss}  proposed in Hard-Net~\cite{Mishchuk2017WorkingHT}. The \emph{hard loss} maximizes the distance between the closest positive and closest negative example in the batch.
Considering the patches sampled from scratch may bring label ambiguity, we improve the \emph{hard loss} by a \emph{neighbor mask}, which makes descriptor training more stable. We formulate description loss as
\begin{equation}
\mathcal{L}_{des}(
\mathbb{D}_{pos},
\mathbb{D}_{ng}
)=
\frac{1}{K}
\sum_{}
max(
0,1+
\mathbb{D}_{pos}-
\mathbb{D}_{ng}
),
\end{equation}
where
\begin{equation}
\mathbb{D}_{pos}(
\mathbf{D}^{k}_{i},
\mathbf{\hat{D}}^{k}_{j}
)
=d(\mathbf{D}^{k}_{i}, \mathbf{\hat{D}}^{k}_{j}),
\end{equation}
and
\begin{equation}
\mathbb{D}_{ng}
=min(
d(\mathbf{D}^{k}_{i}, \mathbf{\hat{D}}^{n}_{j}),
d(\mathbf{D}^{m}_{i}, \mathbf{\hat{D}}^{k}_{j})
).
\end{equation}
Here $\mathbf{\hat{D}}^{n}_{j}$ is the closest non-matching descriptor to $\mathbf{D}^{k}_{i}$ where
\begin{equation}
n=argmin_{n' \neq k}
d(\mathbf{D}^{k}_{i}, \mathbf{\hat{D}}^{n'}_{j})
\;\&\;
E(\mathbf{\hat{p}}^{k}_{j}, \mathbf{\mathbf{p}}^{n'}_{j}) > \mathbf{C}.
\end{equation}
$\mathbf{D}^{m}_{i}$ is the closest non-matching descriptor to $\mathbf{\hat{D}}^{k}_{j}$ where
\begin{equation}
m=argmin_{m' \neq k}
d(\mathbf{D}^{m'}_{i}, \mathbf{\hat{D}}^{k}_{j})
\&
E(\mathbf{p}^{m'}_{i}, \mathbf{p}^{k}_{i}) > \mathbf{C}.
\end{equation}
Function $E$ computes the Euclidean distance between the centroids of the two patches.
We call it a \emph{neighbor mask}.
If a patch $\mathbf{p}^{m'}_{i}$ is very close to $\mathbf{p}^{k}_{i}$, then $\mathbf{p}^{m'}_{i}$ and $\mathbf{\hat{p}}^{k}_{j}$ should be a correct match.
If a patch $\mathbf{\hat{p}}^{n'}_{j}$ is very close to $\mathbf{\hat{p}}^{k}_{j}$, then $\mathbf{\hat{p}}^{n'}_{j}$ and $\mathbf{p}^{k}_{i}$ should be a correct match.
Therefore, we call patch $\mathbf{p}^{m'}_{i}$ a positive patches of $\mathbf{\hat{p}}^{k}_{j}$ if their centroid distance is less than a threshold $\mathbf{C}$.
We mask it when collecting negative samples for  $\mathbf{\hat{p}}^{k}_{j}$.


In summary, we train description network with $\mathcal{L}_{des}$ and train detection network with $ \mathcal{L}_{det} $:
\begin{equation}
\mathcal{L}_{det}=\lambda_{1}\mathcal{L}_{score}+\lambda_{2}\mathcal{L}_{patch}.
\end{equation}

\section{Experiments}

\subsection{Training}
\textbf{Training data}. We trained our network on open dataset \emph{HPatches}~\cite{balntas2017hpatches}. 
This is a recent dataset for local patch descriptor evaluation consists of 116 sequences of 6 images with known homography. 
The dataset is split into two parts: $viewpoint$ - 59 sequences with significant viewpoint change and $illumination$ - 57 sequences with significant illumination change, both natural and artificial. We split the viewpoint sequences by a ratio of $0.9$ (53 sequences for training and validation, and rest 6 sequences for testing).

At training stage, we resized all images into $320\times240$, then converted images to gray for simplicity and normalized them individually using their mean and standard deviation.
Differ with LF-Net~\cite{Ono2018LFNetLL}, we do not have depth maps for each image, so all pixels in the image were used for training.

About the training patches for description extractor, we cropped image patches and resized them to $32\times32$ by selecting the top $K$ keypoints with their orientation and scale. 
To keep differentiability, we used a bilinear sampling scheme of~\cite{Jaderberg2015SpatialTN} for cropping.

\textbf{Training detail}. 
At training stage, we extracted $K=512$ keypoints for training, but at the testing stage, we can choose as many keypoints as desired. 
For optimization, we used ADAM~\cite{Kingma2014AdamAM}, and set initial learning rate $0.1$ both for detector and descriptor, and trained descriptor twice and then trained detector once. The $\mathbf{C}$ in \emph{neighbor mask} is 5.

\begin{table}
	\begin{center}
		\begin{tabular}{|l|c|c|c|}
			\hline 
									& HP-illum		& HP-view 		& EF \\
			\hline
			SIFT 		  			& 0.490			& 0.494			& 0.296 \\
			SURF 		  			& 0.493			& 0.481			& 0.235 \\
			\hline
			L2-Net+DoG    			& 0.403			& 0.394			& 0.189 \\
			L2-Net+SURF   			& 0.627			& 0.629			& 0.307 \\
			L2-Net+FAST   			& 0.571			& 0.431			& 0.229 \\
			L2-Net+ORB    			& 0.705			& 0.673			& 0.298 \\
			L2-Net+Zhang et al.		& 0.685			& 0.425			& 0.235 \\
			\hline
			Hard-Net+DoG   			& 0.436			& 0.468			& 0.206 \\
			Hard-Net+SURF  			& 0.650			& 0.668			& 0.334 \\
			Hard-Net+FAST  			& 0.617			& 0.630			& 0.290 \\
			Hard-Net+ORB   			& 0.616			& 0.632			& 0.238 \\
			Hard-Net+Zhang et al.	& 0.671			& 0.557			& 0.273 \\
			\hline
			LF-Net		  			& 0.617			& 0.566			& 0.251 \\
			RF-Net	      			&\textbf{0.783} &\textbf{0.808} &\textbf{0.453} \\
			\hline
		\end{tabular}
	\end{center}
	\caption{
		\textbf{Comparison to state-of-the-art and baselines.} 
		Average match score measured with three evaluation protocol in each image sequences. 
		Individual descriptors and detector are trained under same sequences as end-to-end networks.
		All feature descriptors is 128 dimension and L2-Normalized.
	}
	\label{tab:matchscore}
\end{table}

\begin{figure*}[t]
	
	\centering
	\includegraphics[width=\textwidth]{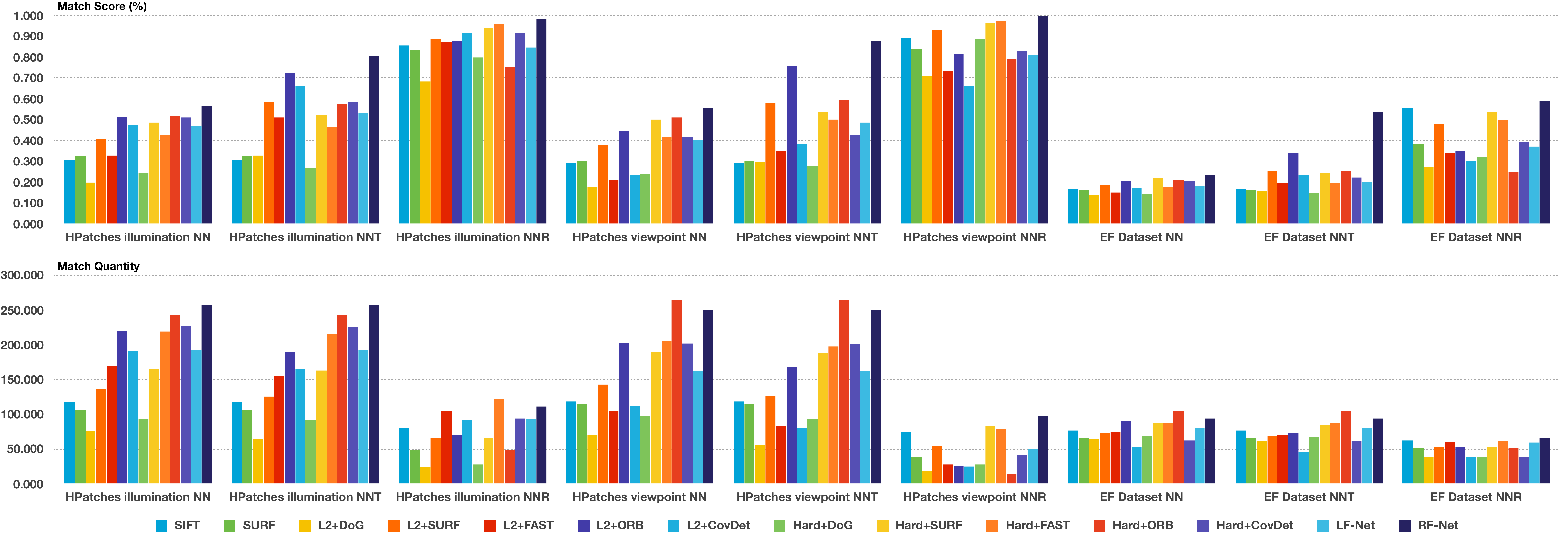}
	
	\caption{
		\textbf{Top}: average match score in each evaluation protocol and sequences. 
		\textbf{Bottom}: average match quantity in each evaluation protocol and sequences. 
		The match score of RF-Net outperforms the closest competitor by a large margin in \emph{EF} Dataset NNT.}
	\label{fig:hist}
\end{figure*}

\subsection{Evaluation data and protocol}
Beside \emph{HPatches} illumination and viewpoint sequences, we also evaluated our model on \emph{EF} Dataset~\cite{Zitnick2011EdgeFI}. 
\emph{EF} Dataset has 5 sequences of 38 images which contains drastic illumination and background clutter changes. 

The definition of a match depends on the matching strategy. To evaluate the entire local feature pipeline performance, we use three matching strategies from~\cite{Mikolajczyk2003APE} to calculate match score for quantitative evaluation:
\begin{itemize}
	\item The first is nearest neighbor (NN) based matching, two regions $\mathbf{A}$ and $\mathbf{B}$ are matched if the descriptor $\mathbf{D_{B}}$ is the nearest neighbor to $\mathbf{D_{A}}$. With this approach, a descriptor has only one match.
	
	\item The second is nearest neighbor with a threshold (NNT) based matching, two regions $\mathbf{A}$ and $\mathbf{B}$ are matched if the descriptor $\mathbf{D_{B}}$ is the nearest neighbor to $\mathbf{D_{A}}$ and if the distance between them is below a threshold $t$.
	
	\item The third is nearest neighbor distance ratio (NNR) based matching, two regions $\mathbf{A}$ and $\mathbf{B}$ are matched if $||\mathbf{D_{A}}-\mathbf{D_{B}}||\;/\;||\mathbf{D_{A}}-\mathbf{D_{C}}||<t$, 
	where $\mathbf{D_{B}}$ is the first and $\mathbf{D_{C}}$ is the second nearest neighbor to $\mathbf{D_{A}}$.
\end{itemize}

All matching strategies compared each descriptor of the reference image with each descriptor of the transformed image. 
To emphasize accurate localization of keypoints, follow~\cite{Ono2018LFNetLL,Rosten2010FasterAB}, we used $ 5 $-pixel threshold instead of the overlap measure used in~\cite{Mikolajczyk2003APE}. 
All learned descriptors have been L2 normalized and their distance range is at $ [0,2] $. 
For fairness, we also L2 normalized hand-craft descriptors and set $ 1.0 $ as the nearest neighbor threshold and $ 0.7 $ as the nearest neighbor distance ratio threshold. 

\subsection{Results on match performance}
We compared RF-Net to three types of methods, the first one is full local feature pipelines, SIFT~\cite{lowe2004distinctive}, SURF~\cite{Bay2008SpeededUpRF}, LF-Net~\cite{Ono2018LFNetLL}. 
The second one is hand-craft detector integrated with learned descriptor, that is DoG~\cite{lowe2004distinctive}, SURF~\cite{Bay2008SpeededUpRF} FAST~\cite{Rosten2010FasterAB} and ORB~\cite{Rublee2011ORBAE} integrated with L2-Net~\cite{Tian2017L2NetDL} and Hard-Net~\cite{Mishchuk2017WorkingHT}.
The third one is learned detector integrated with a learned descriptor, that is Zhang~\emph{et al.}~\cite{Zhang2017LearningDA} integrated with L2-Net~\cite{Tian2017L2NetDL} and Hard-Net~\cite{Mishchuk2017WorkingHT}.
We use the authors' release for L2-Net, Hard-Net, LF-Net and Zhang~\emph{et al.}, and OpenCV for the rest.
For LF-Net and Zhang~\emph{et al.}, we trained them same as RF-Net in 53 viewpoint image sequences cut from \emph{HPatches}~\cite{balntas2017hpatches}. 
For Hard-Net and L2-Net, we trained them in 53 viewpoint patches sequences provided by \emph{HPatches}. 
The length of all feature descriptors is 128 dimension and L2-Normalized.

\begin{table}
	\begin{center}
		\begin{tabular}{|l|c|c|c|}
			\hline 
							& HP-illum		& HP-view 		& EF \\
			\hline
			LF-Det+LF-Des	& 0.617			& 0.566			& 0.251 \\
			RF-Det+LF-Des 	& 0.720			& 0.665			& 0.325 \\
			LF-Det+RF-Des   & 0.744			& 0.714			& 0.361 \\			
			\hline
			RF-Det+RF-Des   & \textbf{0.783}& \textbf{0.808}& \textbf{0.453}\\
			\hline
		\end{tabular}
	\end{center}
	\caption{
		\textbf{Ablation studies.}
		Average match score measured with three evaluation protocol in each image sequences. 
		All methods are trained end-to-end with the same training data.
		LF-Des represents the descriptor used in LF-Net~\cite{Ono2018LFNetLL}, and RF-Des represents the descriptor used in our RF-Net.
		The pipeline performance improved by replacing LF-Det with RF-Det.}
	\label{tab:abla1}
\end{table}

\begin{table}
	\begin{center}
		\begin{tabular}{|l|c|c|c|}
			\hline 
			& HP-illum			& HP-view			& EF \\
			\hline
			RF-Net(No Mask)		& 0.734				& 0.753				& 0.423 \\
			RF-Net(No Orient)	& 0.762				& 0.791				& 0.432 \\
			RF-Net				& \textbf{0.783}	& \textbf{0.808}	& \textbf{0.453} \\
			\hline
		\end{tabular}
	\end{center}
	\caption{
		\textbf{Ablation studies.} 
		Average match score measured with three evaluation protocol in each image sequences. 
		RF-Net(No Mask) means RF-Net trained without the \emph{neighbor mask} loss function item. 
		RF-Net(No Orient) means RF-Net trained without orientation estimation module.
	}
	\label{tab:abla2}
\end{table}

\begin{figure}
	\centering\includegraphics[width=\columnwidth]{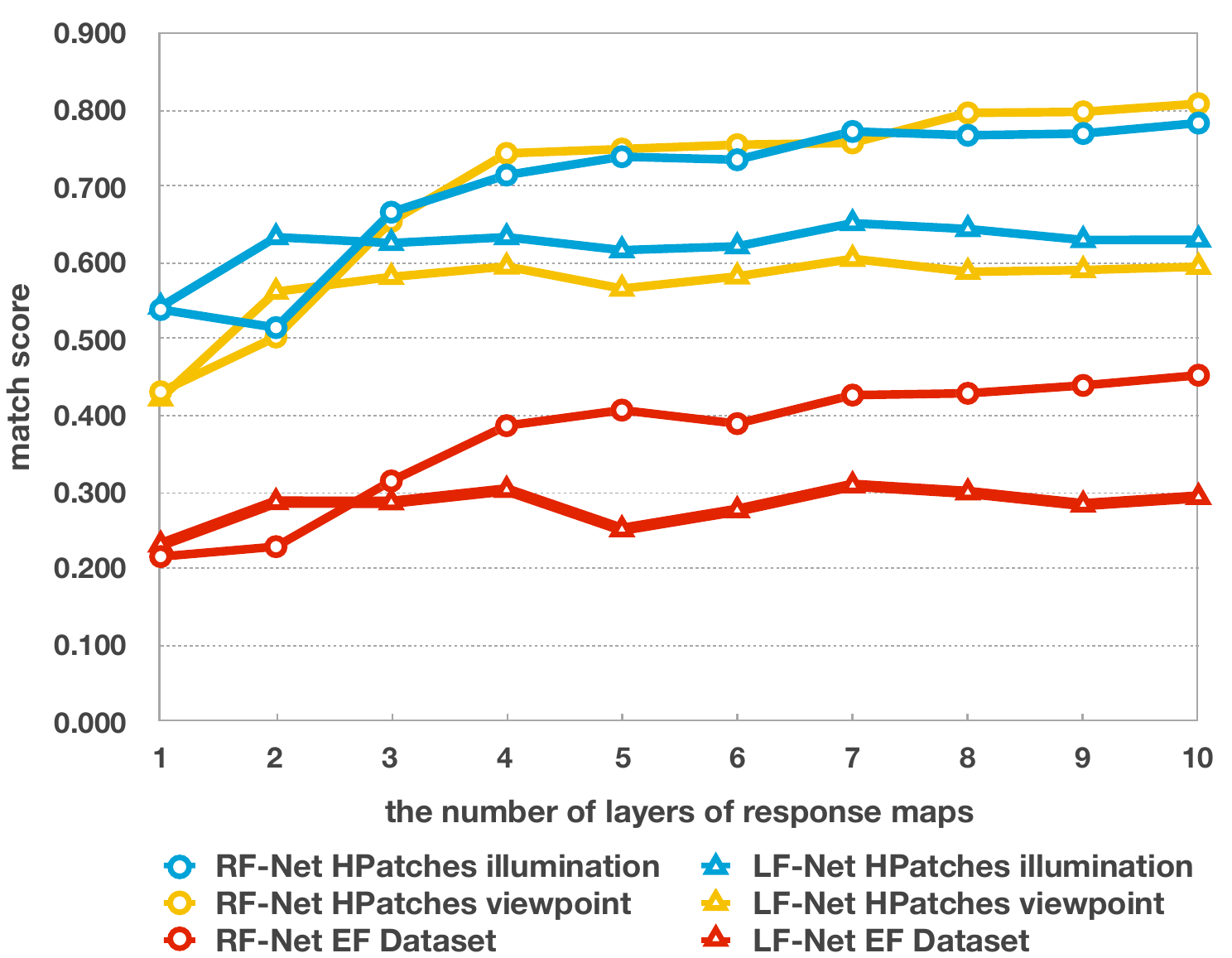}
	\caption{Match score comparison of RF-Net and LF-Net under different $N$-layer response maps.}
	\label{fig:N_exp}
\end{figure}

\begin{table*}
	\begin{center}
		\begin{tabular}{|l|c|c|c|c|c|c|c|c|c|c|}
			\hline 
			& \multicolumn{3}{|c|}{HPatches-illum} & \multicolumn{3}{|c|}{HPatches-view} & \multicolumn{3}{|c|}{EF Dataset} & \\
			\hline  
			\#keypoints & 512     			& 1024   			& 2048   			& 512     			& 1024   			& 2048  			& 512     			& 1024   			& 2048  			& Average \\
			\hline
			DoG    		& 0.638   			& 0.672  			& 0.687  			& 0.609   			& 0.645  			& 0.646 			& 0.512   			& 0.572  			& 0.572 			& 0.617 \\
			FAST   		& 0.790   			& 0.853  			& 0.899  			& \underline{0.704}	& \textbf{0.801}	& \textbf{0.853}	& 0.560   			& \textbf{0.691}	& \textbf{0.791}	& \textbf{0.771} \\
			ORB    		& 0.780   			& 0.830  			& 0.869  			& \textbf{0.709}	& \underline{0.769}	& \underline{0.821}	& 0.524   			& 0.598  			& 0.669 			& 0.730 \\
			SURF   		& 0.708   			& 0.746  			& 0.746  			& 0.633   			& 0.665  			& 0.665 			& 0.569   			& 0.613  			& 0.613 			& 0.662 \\
			Zhang et al.	& \textbf{0.827}	& \textbf{0.894}	& \underline{0.917}	& 0.516   			& 0.664  			& 0.747 			& \textbf{0.588}	& 0.638  			& 0.638 			& 0.714 \\	
			LF-Det 		& 0.727   			& 0.854  			& \textbf{0.922}	& 0.558   			& 0.650  			& 0.717 			& 0.507   			& 0.586  			& 0.667 			& 0.688\\
			RF-Det 		& \underline{0.793}	& \underline{0.868}	& 0.889  			& 0.689				& 0.723  			& 0.729 			& \underline{0.575}	& \underline{0.669}	& \underline{0.704}	& \underline{0.738}\\
			\hline
		\end{tabular}
	\end{center}
	\caption{
		Repeatability at different keypoints in three evaluation sequences.
	}
	\label{tab:reapeat}
\end{table*}

As shown in Table.~\ref{tab:matchscore}, our RF-Net outperforms all others and sets the new state-of-the-art on \emph{HPatches} and \emph{EF} Dataset. Our RF-Net outperforms the closest competitor by $ 11\% $, $ 20\% $ and $ 35\% $ relative in the three sequences.

Match score represents the correct ratio in method prediction, while match quantity represents the correct predicted quantity. 
Figure.~\ref{fig:hist} depicts the match score and match quantity in all evaluations, and our RF-Net get both high match score and match quantity.
The pipeline of ORB combined with Hard-Net also achieves good match quantity in NN and NNT protocols, but it does not perform well in NNR protocol. 
This indicates descriptors extracted by this pipeline have high nearest neighbor distance ratio, while our RF-Net does not have this problem.

\begin{figure}[h!bt]	
	\centering
	\begin{tabular}{ccc}
		\includegraphics[width=0.14\textwidth]{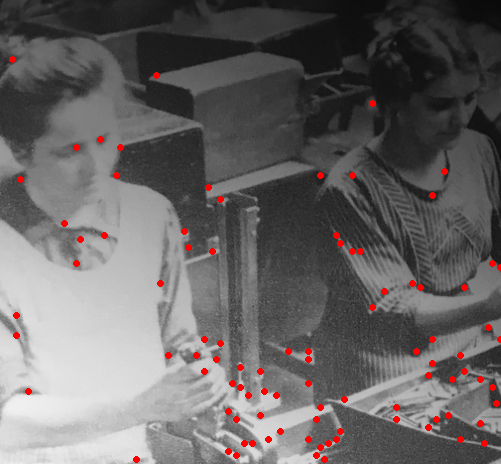} &
		\includegraphics[width=0.14\textwidth]{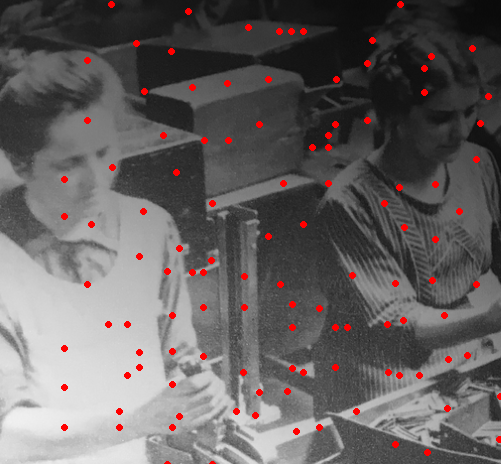} &
		\includegraphics[width=0.14\textwidth]{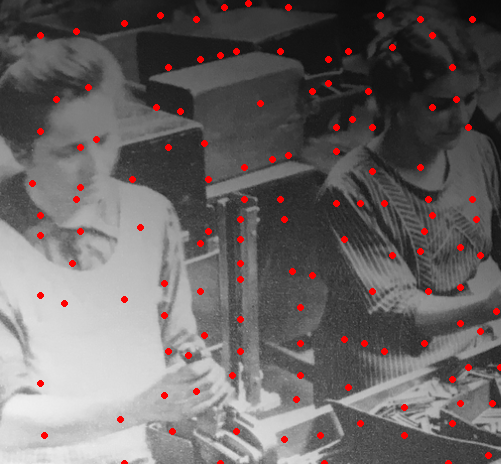} \\
		\includegraphics[width=0.14\textwidth]{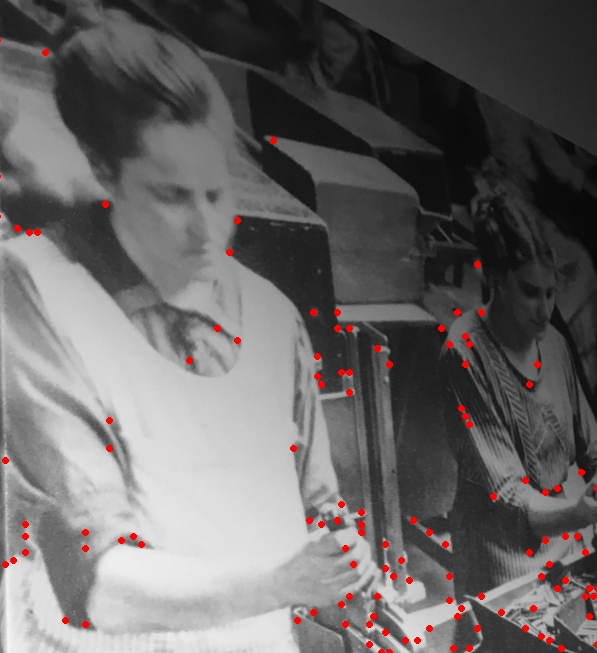} &
		\includegraphics[width=0.14\textwidth]{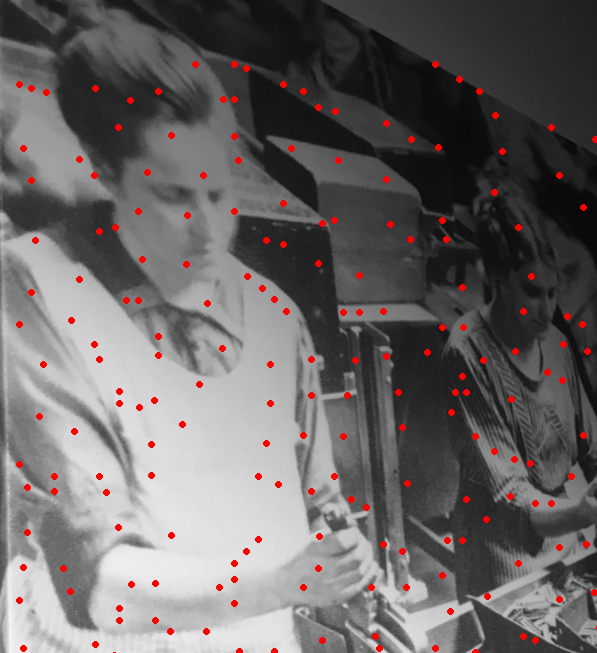} &
		\includegraphics[width=0.14\textwidth]{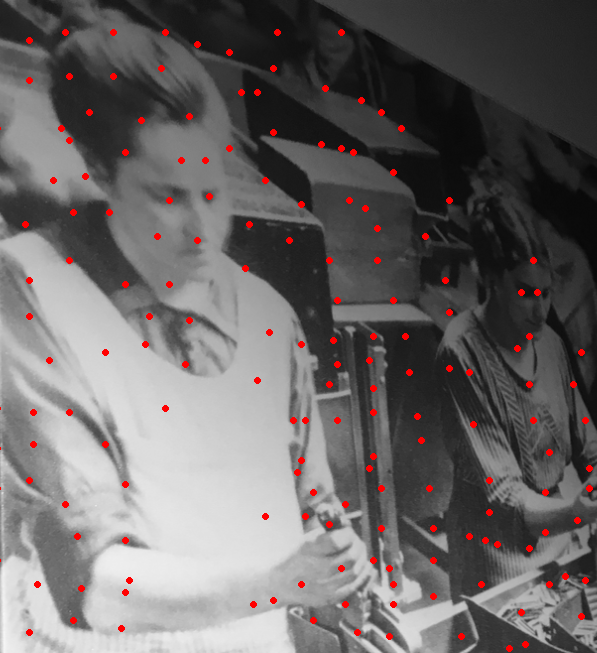} \\
		(a) FAST & (b) LF-Det & (c) RF-Det
	\end{tabular}
	\caption{
		Visualize keypoints detected by FAST, LF-Det and RF-Det. 
		Keypoints detected by RF-Det and LF-Det are more sparse than FAST. 
	}
	\label{fig:kpvis}
\end{figure}

We also give the experiment results about how $ N $ response layers effect the RF-Net and LF-Net in Figure.~\ref{fig:N_exp}.
For RF-Net, match score increases with the number of response layers and saturates after $ N=8 $,
and the gap in performance between the LF-Net and RF-Net starts from $ N=3 $ and increases as $ N $ increases.
This demonstrates that receptive field based response maps are more effective than abstract feature based method

\subsection{Discussions and ablation studies}
In this section, we examine the importance of various components of our architecture.
We replaced LF-Det with RF-Det, and trained them with the same training data to show the effectiveness of our RF-Det. 
Table.~\ref{tab:abla1} shows the pipeline performance improved by replacing LF-Det with RF-Det.

To mine the effectiveness of modules in RF-Net, we try to remove \emph{neighbor mask} and orientation estimation module from RF-Net. 
Table.~\ref{tab:abla2} shows \emph{neighbor mask} brings remarkable match improvement to RF-Net. 
Even we removed orientation prediction, our RF-Net still gets state-of-the-art match score, this represents the robustness of our RF-Det.

\subsection{Results on repeatability}
Table.~\ref{tab:reapeat} shows the repeatability performance of hand-craft approaches, Zhang~\emph{et al.}, LF-Det and our RF-Det. 
Although FAST does not perform best on image match, it gets the highest repeatability.
Pipeline of matching is a cooperation task between detector and descriptor.
As shown in Figure.~\ref{fig:kpvis}, the keypoints detected by learned end-to-end detector (LF-Det and RF-Det) are more sparse than FAST.
This indicates sparse keypoints are easier to match, because too close keypoints may produce patches too similar to match.
Therefore, a parse detector works better on this task.
Compare RF-Det with LF-Det, RF-Det indeed gets a higher repeatability than LF-Det in all sequences. 
This also benefited from the receptive field design.

\subsection{Qualitative results}
In Figure.~\ref{fig:qualitative}, We also give some qualitative results on the task of matching challenging pairs of images provided by \emph{EF} Dataset and \emph{HPatches}. 
We selected top $K=1024$ keypoints firstly, then matched them by the nearest neighbor distance ratio matching strategy with $0.7$ threshold. We compared our method with the SIFT~\cite{lowe2004distinctive}, FAST~\cite{Rosten2010FasterAB} detector integrated with Hard-Net~\cite{Mishchuk2017WorkingHT}, and LF-Net~\cite{Ono2018LFNetLL}. 
The images in top two rows are from \emph{EF} Dataset, and the images in bottom two rows are from \emph{HPatches}.
These images are under large illumination changes or perspective transformation.
As shown in Figure.~\ref{fig:qualitative}, our method produced the maximum quantity of green matching lines and fewer red failed match lines.

\section{Conclusions}
We present a novel end-to-end deep network, RF-Net, for local feature detection and description. 
To learn more robust response maps, we propose a novel keypoint detector based on receptive field. 
We also design a loss function term, \emph{neighbor mask}, to learn a more stable descriptor.
Both of these designs bring significant performance improvement to the matching pipeline. 
We conducted qualitative and quantitative evaluations in three data sequences and showed significant improvements over existing state-of-the-art.

\section*{Acknowledgments}
This work was supported by the National Natural Science Foundation of China (No. U1605254, 61728206) and the National Science Foundation of USA EAR-1760582.\\

{\small
\bibliographystyle{ieee}
\bibliography{egbib}
}

\end{document}